%% file: main.tex
\documentclass[11pt, a4paper, logo, onecolumn, copyright]{googledeepmind} %

\pdfoutput=1

\usepackage{CJKutf8}
\usepackage[numbers]{natbib}
\usepackage{xspace}
\usepackage{algorithmic}
\usepackage{algorithm}
\usepackage{xcolor}
\usepackage{float}
\usepackage{booktabs}
\usepackage{wrapfig}
\usepackage{subcaption}
\usepackage{graphicx}
\usepackage{sidecap}  %

\usepackage{amsmath}
\usepackage{amssymb}
\usepackage{mathtools}
\usepackage{amsthm}
\usepackage{enumitem}

\usepackage{nidanfloat}

\newcommand{\eg}{e.g., }

\newcommand{\algname}{PIVOT\xspace}
\newcommand{\algfullname}{Prompting with Iterative Visual Optimization\xspace}

\date{January 2024}

\title{\algname: Iterative Visual Prompting for VLMs\\with Applications to Zero-Shot Robotic Control}
\title{\algname: Iterative Visual Prompting Elicits Actionable Knowledge for VLMs}

\author{Soroush Nasiriany$^{*,\dagger,1,3}$, Fei Xia$^{*,1}$, Wenhao Yu$^{*,1}$, Ted Xiao$^{*,1}$, Jacky Liang$^{1}$, Ishita Dasgupta$^{1}$, Annie Xie$^{2}$, Danny Driess$^{1}$, Ayzaan Wahid$^{1}$, Zhuo Xu$^{1}$, Quan Vuong$^{1}$, Tingnan Zhang$^{1}$, Tsang-Wei Edward Lee$^{1}$, Kuang-Huei Lee$^{1}$, Peng Xu$^{1}$, Sean Kirmani$^{1}$, Yuke Zhu$^{3}$, Andy Zeng$^{1}$, Karol Hausman$^{1}$, Nicolas Heess$^{1}$, Chelsea Finn$^{1}$, Sergey Levine$^{1}$, Brian Ichter$^{*,1}$ \\
$^1$Google DeepMind, $^2$Stanford University, $^3$The University of Texas at Austin \\
Correspond to: \texttt{\{soroushn, xiafei, magicmelon, tedxiao, ichter\}@google.com}\\
{Website: \href{http://pivot-prompt.github.io}{pivot-prompt.github.io}} and
HuggingFace: \href{https://huggingface.co/spaces/pivot-prompt/pivot-prompt-demo}{https://huggingface.co/spaces/pivot-prompt/pivot-prompt-demo}
}

\begin{abstract}
Vision language models (VLMs) have shown impressive capabilities across a variety of tasks, from logical reasoning to visual understanding.
This opens the door to richer interaction with the world, for example robotic control. However, VLMs produce only textual outputs, while robotic control and other spatial tasks require outputting continuous coordinates, actions, or trajectories. How can we enable VLMs to handle such settings without fine-tuning on task-specific data?
\vspace{0.1cm}

In this paper, we propose a novel visual prompting approach for VLMs that we call \algfullname (\algname),
which casts tasks as iterative visual question answering. In each iteration, the image is annotated with a visual representation of proposals that the VLM can refer to (e.g., candidate robot actions, localizations, or trajectories). The VLM then selects the best ones for the task. 
These proposals are iteratively refined, allowing the VLM to eventually zero in on the best available answer.
We investigate \algname on real-world robotic navigation, real-world manipulation from images, instruction following in simulation, and additional spatial inference tasks such as localization.
We find, perhaps surprisingly, that our approach enables zero-shot control of robotic systems without any robot training data, navigation in a variety of environments, and other capabilities. 
Although current performance is far from perfect, our work highlights potentials and limitations of this new regime and shows a promising approach for Internet-Scale VLMs in robotic and spatial reasoning domains.

\end{abstract}

\begin{document}
\maketitle

\begin{figure}[htb]
    \centering
    \includegraphics[width=0.86\linewidth]{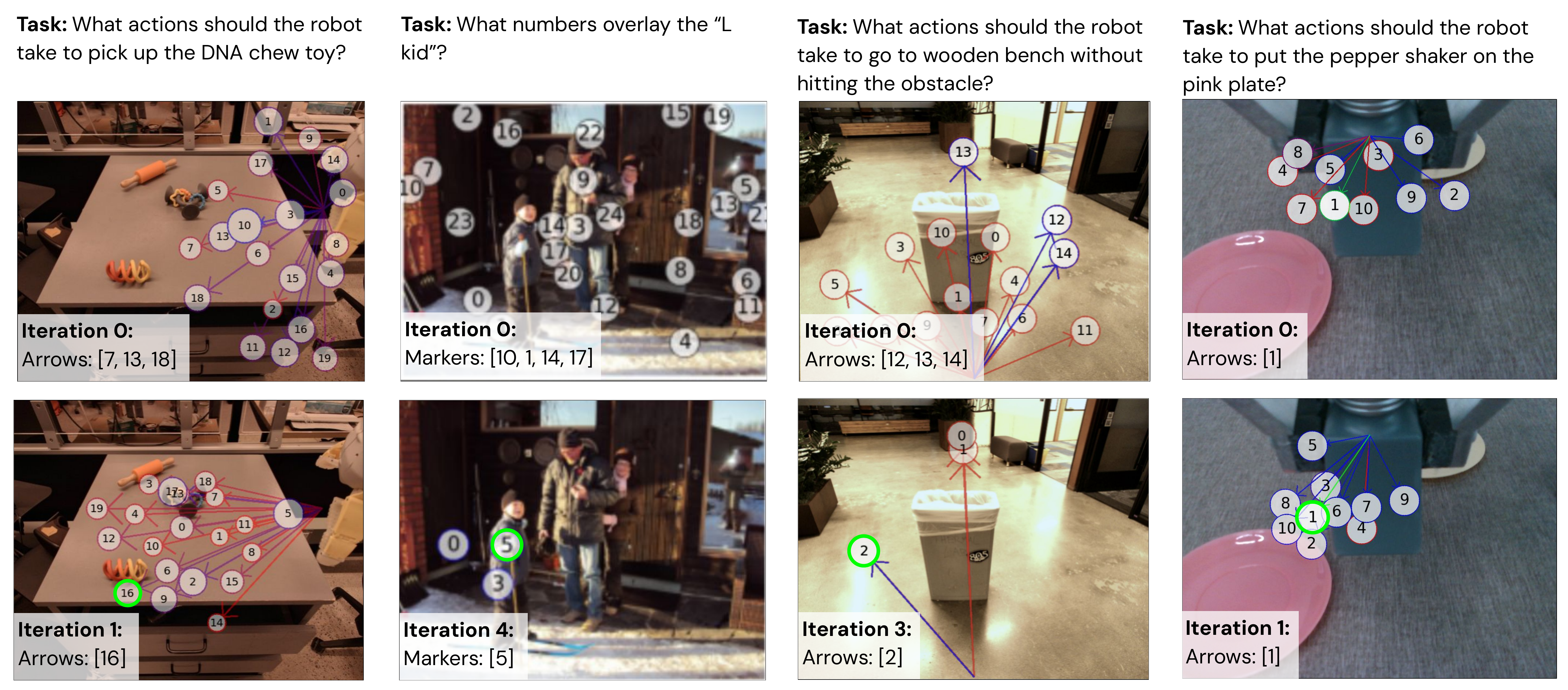}
    \caption{\small \algfullname (\algname) casts spatial reasoning tasks, such as robotic control, as a VQA problem. This is done by first annotating an image with a visual representation of robot actions or 3D coordinates, then querying a VLM to select the most promising annotated actions seen in the image. The best action is iteratively refined by fitting a distribution to the selected actions and requerying the VLM. This procedure enables us to solve complex tasks that require outputting grounded continuous coordinates or robot actions utilizing a VLM without any domain-specific training.}
    \label{fig:teaser}
\end{figure}

\section{Introduction}

Large language models (LLMs) have shown themselves capable of solving a broad range of practical problems, from code generation to question answering and even logical deduction~\cite{brown2020language, austin2021program,wei2022chain}. The extension of LLMs to multi-modal inputs, resulting in powerful vision-language models (VLMs), enables models that handle much richer visual modalities~\cite{team2023gemini,openai2023gpt4v,alayrac2022flamingo,chen2023pali}, which makes it feasible to interact not only with natural language but directly with the physical world.
However, most VLMs still only \emph{output} textual answers, seemingly limiting such interactions to high-level question answering. Many real-world problems are inherently spatial: controlling the trajectory of a robotic arm, selecting a waypoint for a mobile robot, choosing how to rearrange objects on a table, or even localizing keypoints in an image. Can VLMs be adapted to solve these kinds of embodied, physical, and spatial problems? And can they do so zero shot, without additional in-domain training data? In this work, we propose an iterative prompting method to make this possible and study the limits and potentials for zero-shot robotic control and spatial inference with VLMs.

Our proposed method is based on a simple insight: although VLMs struggle to produce precise spatial outputs directly,
they can readily select among a discrete set of coarse choices, and this in turn can be used to \emph{refine} this set to provide more precise choices at the next iteration. At each iteration of our iterative procedure, we annotate the image with candidate proposals (i.e., numbered keypoints as in~\citet{yang2023set})
drawn from a proposal distribution, and ask the VLM to rank the degree to which they perform the desired task. We then \emph{refine} this proposal distribution, generate new candidate proposals that are clustered around better regions of the output space, and repeat this procedure. With this optimization approach, the entire loop can be viewed as an iterative optimization similar to the cross-entropy method~\cite{de2005tutorial},
with each step being framed as a visual question compatible with current VLMs without any additional training.
In Figure~\ref{fig:teaser} and throughout this work, we use robot control as a running example, wherein candidates are numbered arrows.

Equipped with our method for extracting spatial outputs from VLMs, we study the limits and potentials of zero-shot VLM inference in a range of domains: robotic navigation, grasping and rearranging objects, language instructions in a simulated robotic benchmark, and non-robot spatial inference through keypoint localization. 
It is important to note that in all of these domains, we use state-of-the-art vision language models, namely GPT-4~\citep{openai2023gpt4v} and Gemini~\citep{gemini2023gemini}, \emph{without any modification or finetuning}. 
Our aim is not necessarily to develop the best possible robotic control or keypoint localization technique, but to study the limits and potentials of such models. 
We expect that future improvements to VLMs will lead to further quantitative gains on the actual tasks. 
The zero-shot performance of VLMs in these settings is far from perfect, but the ability to control robots in zero shot without \emph{any} robotic data, complex prompt design, code generation, or other specialized tools provides a very flexible and general way to obtain highly generalizable systems.

Our main contribution is thus an approach for visual prompting and iterative optimization with VLMs, applications to low-level robotic control and other spatial tasks, and an empirical analysis of potentials and limitations of VLMs for such zero-shot spatial inference.
We apply our approach to a variety of robotic systems and general visually-grounded visual question and answer tasks, and evaluates the kinds of situations where this approach succeeds and fails.
While our current results are naturally specific to current state-of-the-art VLMs, we find that performance improves with larger, more performant VLMs. Thus, as VLM capabilities continue to improve with time, we expect our proposed approach to improve in turn.

\section{Related Work}
\textbf{Visual annotations with VLMs.}
With the increasing capabilities of VLMs, there has been growing interest in understanding their abilities to understand visual annotations \cite{yang2023dawn, shtedritski2023does, yan2023gpt, zheng2024gpt}, improving such capabilities \cite{cai2023making, xu2023pixel}, as well as leveraging them for perception or decision-making tasks \cite{gu2023rt, yang2023set, wen2023road, koh2024visualwebarena, liu20233daxiesprompts}. \citet{shtedritski2023does} identify that VLMs like CLIP \cite{radford2021learning} can recognize certain visual annotations. \citet{yang2023dawn} perform a more comprehensive analysis on the GPT-4 model and demonstrate its ability to understand complex visual annotations.
\citet{yang2023set} demonstrates how such a model can solve visual reasoning tasks by annotating the input image with object masks and numbers.
Several works too have applied visual prompting methods to web navigation tasks~\cite{koh2024visualwebarena,yan2023gpt,zheng2024gpt}, obtaining impressive-zero shot performance.
Our work builds upon these works: instead of taking proposals as given or generating the proposals with a separate perception systems, \algname generates proposals randomly, but then adapt the distribution through iterative refinement. 
As a result, we can obtain relatively precise outputs through multiple iterations, and do not require any separate perception system or any other model at all besides the VLM itself.

\textbf{Prompt optimization.}
The emergence of few-shot in context learning within LLMs~\cite{brown2020language} has lead to many breakthroughs in prompting.
Naturally prompt optimization has emerged as a promising approach, whether with gradients~\cite{li2021prefix,lester2021power} or without gradients, e.g., with human engineering~\cite{kojima2022large} or through automatic optimization in language space~\cite{zhou2022large}.
These automatic approaches are most related to our work and have shown that
language-model feedback~\cite{pryzant2023automatic}, 
answer scores~\cite{zhou2022large,yang2023large,xu2022gps},
and environment feedback~\cite{wang2023voyager} can significantly improve the outputs of LLMs and VLMs.
A major difference between these prior methods and ours is that our iterative prompting uses refinement of the \emph{visual} input, by changing the visual annotations across refinement steps. We optimize prompts ``online'' for a specific query rather than offline to identify a fixed prompt, and show that our iterative procedure leads to more precise spatial outputs.

\textbf{Foundation models for robot reasoning and control.}
In recent years, foundation models have shown impressive results in
robotics from high-level reasoning to low-level control~\citep{firoozi2023foundation,hu2023toward}.
Many early works investigated robotic reasoning and planning regimes where LLMs and language outputs are well suited~\cite{huang2022language,zeng2022socratic,ahn2022can,huang2022inner,liu2023reflect, raman2022planning,silver2023generalized,liu2023llm+,lin2023text2motion,wang2023describe,chen2024spatialvlm}.
To apply foundation models to control tasks, several promising approaches have emerged.
One line of work has shown that foundation-model-selected subgoals are an effective abstraction to feed into policies for navigation~\cite{dorbala2022clip,shah2023lm,chen2023open,huang2023visual,shah2023navigation,gadre2023cows} and manipulation~\cite{cui2022can,shridhar2022cliport}.
Another abstraction that has been shown to be effective for control is LLM generated rewards, which can be optimized within simulation~\cite{huang2023voxposer,yu2023language,ma2023eureka}.
Others have investigated code writing LLMs to directly write code that can be executed via control and perceptive primitives~\cite{liang2023code,singh2023progprompt,wu2023tidybot}.
On simple domains, even few-shot prompting language models has been shown to be capable of control~\cite{mirchandani2023large,wang2023prompt}, while finetuned foundation models have yielded significantly more capable VLM-based controllers~\cite{brohan2023rt,shridhar2022cliport,jiang2022vima,reed2022generalist,gao2023physically,padalkar2023open}.
Unlike these works, we show how VLMs can be applied \textit{zero-shot} to low-level control of multiple real robot platforms.%

\section{\algfullname}

The type of tasks this work considers have to be solved by producing a value $a\in\mathcal{A}$ from a set $\mathcal{A}$ given a task description in natural language $\ell \in \mathcal{L}$ and an image observation $I \in \mathbb{R}^{H \times W \times 3}$.
This set $\mathcal{A}$ can, for example, include continuous coordinates, 3D spatial locations, robot control actions, or trajectories.
When $\mathcal{A}$ is the set of robot actions, this amounts to finding a policy $\pi(\cdot | \ell, I)$ that emits an action $a \in \mathcal{A}$.
The majority of our experiments focus on finding a control policy for robot actions.
Therefore, in the following, we present our method of \algname with this use-case in mind.
However, \algname is a general algorithm to generate (continuous) outputs from a VLM.

\subsection{Grounding VLMs to Robot Actions through Image Annotations}
We propose framing the problem of creating a policy $\pi$ as a Visual Question Answering (VQA) problem.
The class of VLMs we use in this work take as input an image $I$ and a textual prefix $w_p$ from which they generate a distribution $P_\text{VLM}(\cdot|w_p, I)$ of textual completions.
Utilizing this interface to derive a policy raises the challenge of how an action from a (continuous) space $\mathcal{A}$ can be represented as a textual completion.

The core idea of this work is to lift low-level actions into the \emph{visual language} of a VLM, i.e., a combination of images and text, such that it is closer to the training distribution of general vision-language tasks.
To achieve this, we propose the \emph{visual prompt mapping}
\begin{align}
    \big(\hat{I}, w_{1:M}\big) = \Omega(I, a_{1:M}) \label{eq:actionMapping}
\end{align}
that transforms an image observation $I$ and set of candidate actions $a_{1:M}$, $a_j\in\mathcal{A}$ into an annotated image $\hat{I}$ and their corresponding textual labels $w_{1:M}$ where $w_j$ refers to the annotation representing $a_j$ in the image space.
For example, as visualized in Fig.~\ref{fig:teaser}, utilizing the camera matrices, we can project a 3D location into the image space, and draw a visual marker at this projected location.
Labeling this marker with a textual reference, e.g., a number, consequently enables the VLM to not only be queried in its natural input space, namely images and text, but also to refer to spatial concepts in its natural output space by producing text that references the marker labels.
In Section~\ref{sec:exp_ablations} we investigate different choices of the mapping \eqref{eq:actionMapping} and ablate its influence on performance.

\subsection{\algfullname}
\begin{figure}[tbh]
    \centering
    \includegraphics[width=\linewidth]{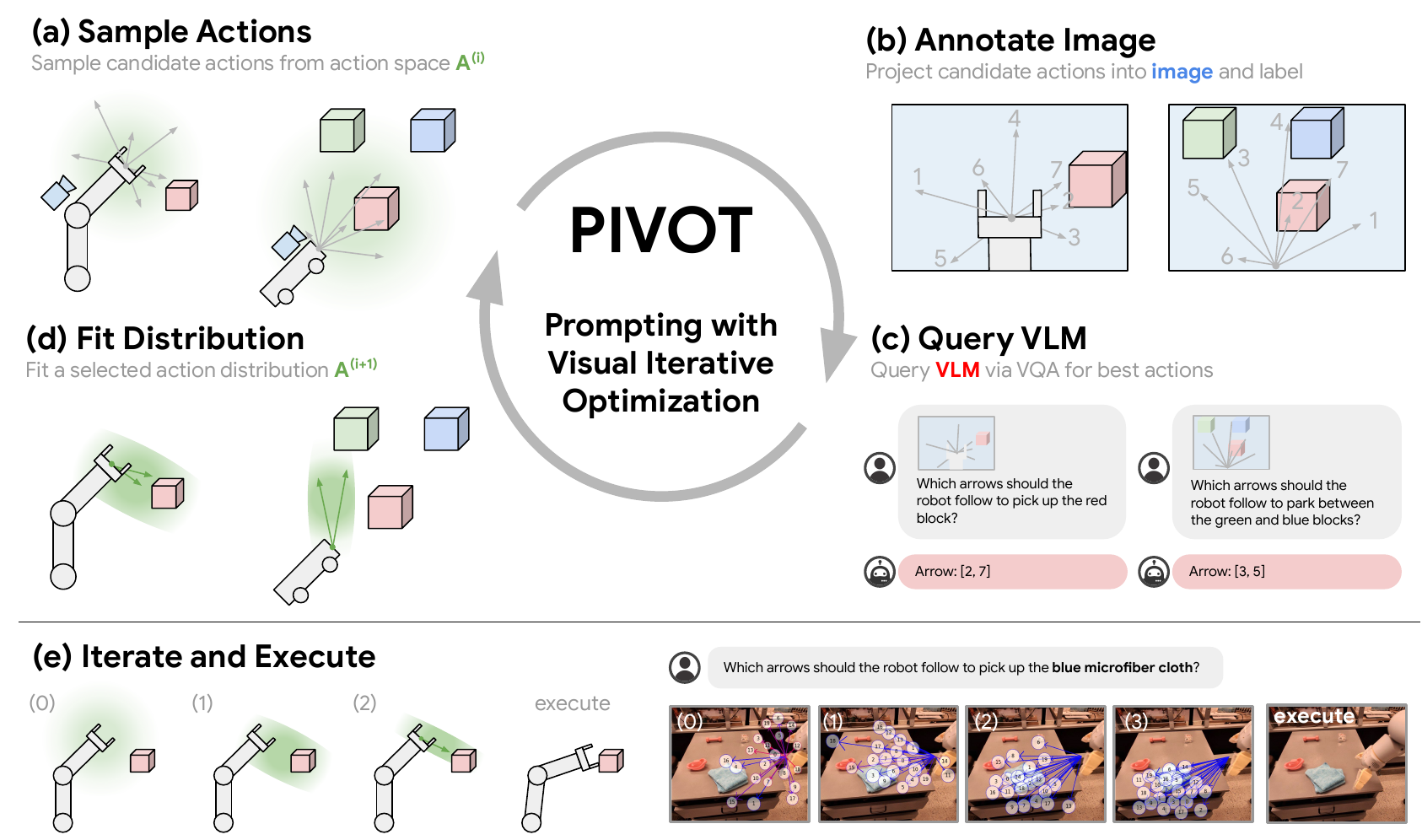}
    \caption{\small \algfullname produces a robot control policy by iteratively (a) sampling actions from an action distribution $\mathcal{A}^{(i)}$, (b) projecting them into the image space and annotating each sample, (c) querying a VLM for the best actions, and (d) fitting a distribution to the selected actions to form $\mathcal{A}^{(i+1)}$. (e) After a set number of iterations, a selected best action is executed.}
    \label{fig:alg}
\end{figure}

Representing (continuous) robot actions and spatial concepts in image space with their associated textual labels allows us to query the VLM $P_\text{VLM}$ to judge if an action would be promising in solving the task.
Therefore, we can view obtaining a policy $\pi$ as solving the optimization problem
\begin{align}
    \max_{a\in\mathcal{A}, w} ~P_\text{VLM}\big(w~\big|~\hat{I}, \ell\big) ~~~~ \text{s.t.}~~~ \big(\hat{I}, w\big) = \Omega(I, a). \label{eq:optProblem}
\end{align}
Intuitively, we aim to find an action $a$ for which the VLM would choose the corresponding label $w$ after applying the mapping $\Omega$.
In order to solve \eqref{eq:optProblem}, we propose an iterative algorithm, which we refer to as \algfullname. %
In each iteration $i$ the algorithm first samples a set of candidate actions $a_{1:M}^{(i)}$ from  a distribution $P_{\mathcal{A}^{(i)}}$ (Figure~\ref{fig:alg} (a)).
These candidate actions are then mapped onto the image $I$ producing the annotated image $\hat{I}^{(i)}$ and the associated action labels $w_{1:M}^{(i)}$ (Figure~\ref{fig:alg} (b)). We then query the VLM on a multiple choice-style question on the labels $w_{1:M}^{(i)}$ to choose which of the candidate actions are most promising (Figure~\ref{fig:alg} (c)).
This leads to set of best actions to which we fit a new distribution $P_{\mathcal{A}^{(i+1)}}$ (Figure~\ref{fig:alg} (d)).
The process is repeated until convergence or a maximum number of steps $N$ is reached.
Algorithm~\ref{alg:pivot} and Figure~\ref{fig:alg} visualize this process.

\subsection{Robust \algname with Parallel Calls}
VLMs can make mistakes, causing \algname to select actions in sub-optimal regions. To improve the robustness of \algname, we use a parallel call strategy, where we first execute $E$ parallel \algname instances and obtain $E$ candidate actions. We then aggregate the selected candidates to identify the final action output. To aggregate the candidate actions from different \algname instances, we compare two approaches: 1) we fit a new action distribution from the $E$ action candidates and return the fitted action distribution, 2) we query the VLM again to select the single best action from the $E$ actions. We find that by adopting parallel calls we can effectively improve the robustness of \algname and mitigate local minima in the optimization process. 

\begin{algorithm}[H]
   \caption{\algfullname}\label{alg:pivot}
\begin{algorithmic}[1]
   \STATE \textbf{Given:} image $I$, instruction $\ell$, action space $\mathcal{A}$, max iterations $N$, number of samples $M$
  \STATE \textbf{Initialize:} $\mathcal{A}^{(0)} = \mathcal{A}$, $i= 0$
  \WHILE{$i < N$}
  \STATE Sample actions $a_{1:M}$ from $P_{\mathcal{A}^{(i)}}$
  \STATE Project actions into image space and textual labels $\big(\hat{I}, w_{1:M}\big)=\Omega(I, a_{1:M})$
  \STATE Query VLM $P_\text{VLM}\big(w~\big|~\hat{I}, \ell\big)$ to determine the most promising actions
  \STATE Fit distribution $P_{\mathcal{A}^{(i+1)}}$ to best actions
  \STATE Increment iterations $i \leftarrow i + 1$
  \ENDWHILE
  \STATE \textbf{Return:} an action from the VLM best actions
\end{algorithmic}
\end{algorithm}

\subsection{\algname Implementation}
Our approach can be used to query the VLM for any type of answer as long as multiple answers can be simultaneously visualized on the image.
As visualized in Figure~\ref{fig:teaser}, for the visual prompting mapping $\Omega$, we represent actions as arrows emanating from the robot or the center of the image if the embodiment is not visible.
For 3D problems, the colors of the arrows and size of the labels indicate forward and backwards movement.
We label these actions with a number label circled at the end of the arrow.
Unless otherwise noted, the VLM used herein was GPT-4V~\cite{openai2023gpt4v}.
For creating the text prompt $w_p$, we prompt the VLM to use chain of thought to reason through the problem and then summarize the top few labels.
The distributions $P_{\mathcal{A}}$ in Algorithm~\ref{alg:pivot} are approximated as isotropic Gaussians.

\section{Experiments}\label{sec:exp}

We investigate the capabilities and limitations of \algname for visuomotor robotic control and visually grounded (\eg spatial) VQA. Our primary examples involve action selection for control because (a) it requires fine-grained visual grounding, (b) actions can be difficult to express in language, and (c) it is often bottlenecked by visual generalization, which benefits from the knowledge stored within pre-trained VLMs.
We aim to understand both the strength and weaknesses of our approach, and believe that (i) identifying these limitations and (ii) understanding how they may be alleviated via scaling and by improving the underlying foundation models are main contributions of this work.
Specifically, we seek to answer the questions:
\begin{enumerate}[leftmargin=15pt,noitemsep,nolistsep]
    \item How does \algname perform on robotic control tasks?
    \item How does \algname perform on object reference tasks?
    \item What is the influence of the different components of \algname (textual prompting, visual prompting, and iterative optimization) on performance?
    \item What are the limitations of \algname with current VLMs?
    \item How does \algname scale with VLM performance?
\end{enumerate}

\subsection{Robotics Experimental Setup}\label{sec:exp_performance}

We evaluate \algname across the following robot embodiments, which are visualized in Figure~\ref{fig:embodiments} and described in detail in Appendix~\ref{app:robots}:
\begin{itemize}[leftmargin=15pt,noitemsep,nolistsep]
    \item Mobile manipulator with a head-mounted camera for both navigation (2D action space, Figure~\ref{fig:embodiments} (a) and manipulation tasks (4D end-effector relative Cartesian $(x, y, z)$ and binary gripper action space, Figure~\ref{fig:embodiments} (b).
    \item Franka arm with a wrist-mounted camera and a 4D action space (end-effector relative Cartesian $(x, y, z)$ and gripper). Results shown in Appendix~\ref{app:franka}.
    \item RAVENS~\cite{zeng2021transporter} simulator, with an overhead camera and a pick and place pixel action space. Results shown in Appendix~\ref{app:ravens}. %
\end{itemize}

\begin{figure}[H]
     \centering
     \includegraphics[width=0.8\linewidth]{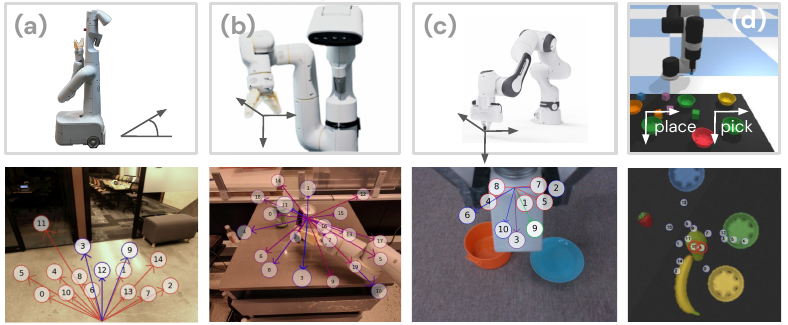}
        \caption{\small We evaluate \algname on several robot embodiments including: a mobile manipulator for (a) navigation  and (b) manipulation, (c) single Franka arm manipulation, and (d) tabletop pick-and-place \cite{zeng2021transporter}.}
        \label{fig:embodiments}
\end{figure}

\subsection{Zero-shot Robotic Control in the Real World}\label{sec:exp_on_robot}

\begin{figure}[htb]
     \centering
     \begin{subfigure}[b]{0.29\textwidth}
         \centering
         \includegraphics[width=\textwidth]{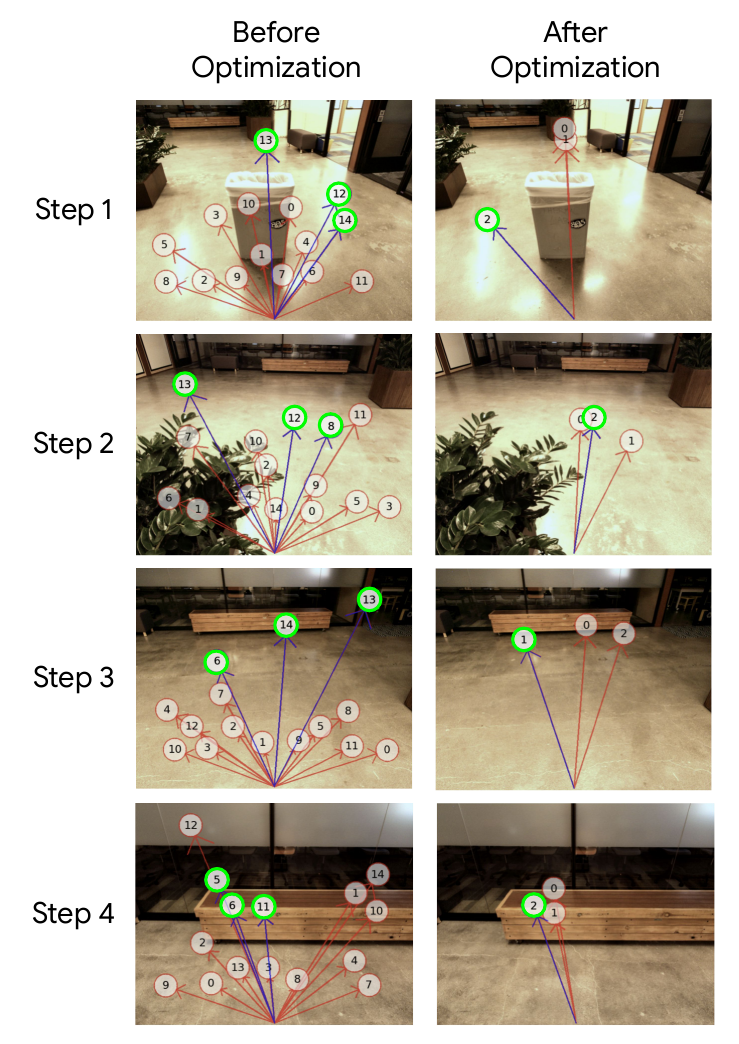}
         \caption{Navigation: ``Help me find a place to sit and write''}
         \label{fig:rollout_meta_nav}
     \end{subfigure}
     \begin{subfigure}[b]{0.28\textwidth}
         \centering
         \includegraphics[width=\textwidth]{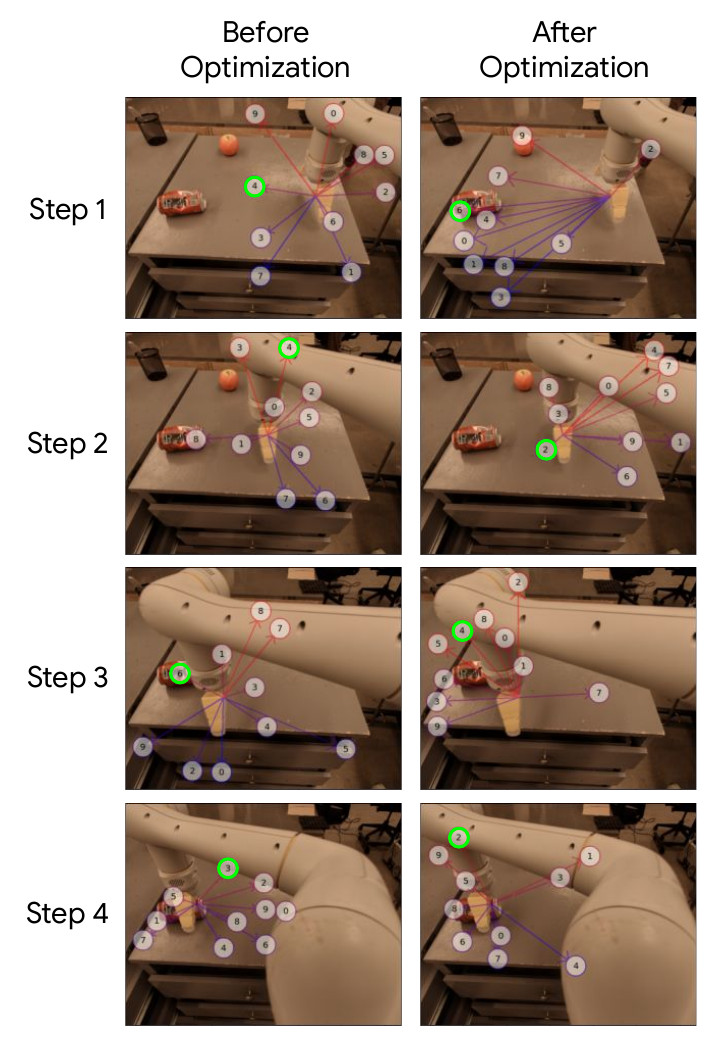}
         \caption{Manipulation: ``Pick up the coke can''}
         \label{fig:rollout_meta_manip}
     \end{subfigure}
     \begin{subfigure}[b]{0.355\textwidth}
         \centering
         \includegraphics[width=\textwidth]{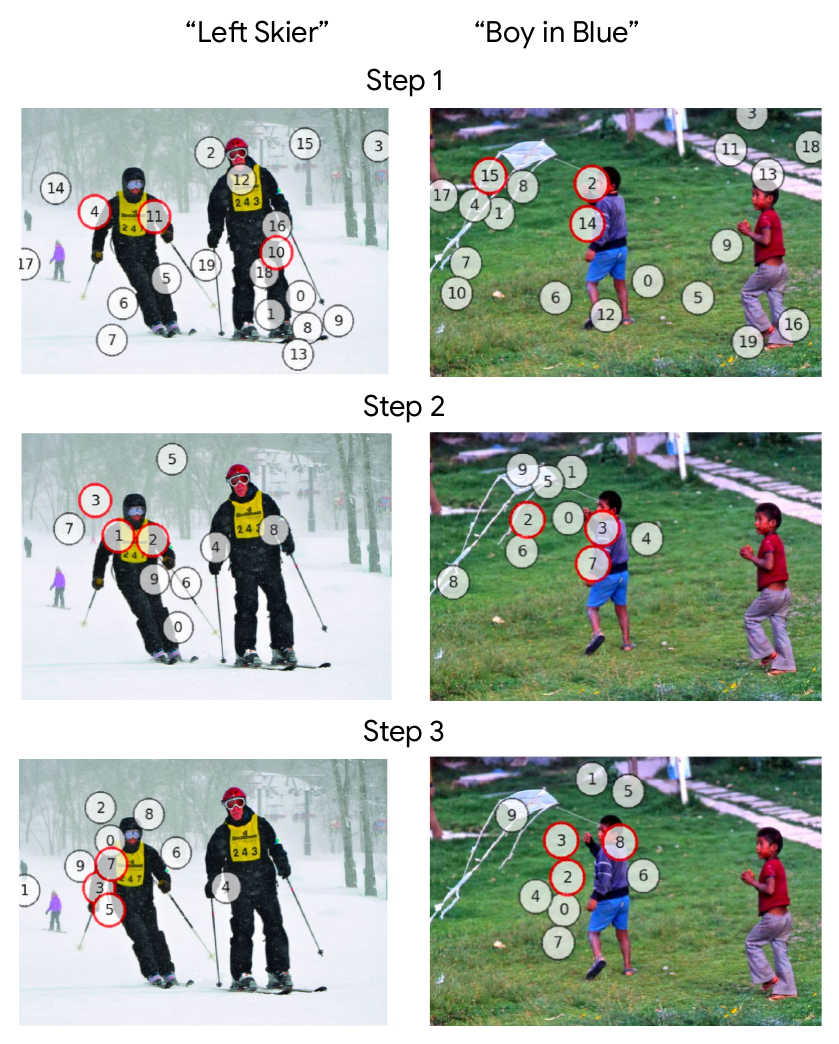}
         \caption{RefCOCO spatial reasoning}
         \label{fig:rollout_ref_coco}
     \end{subfigure}
     \caption{\small (a) An example rollout on a real-world navigation task. We use three parallel calls to generate samples. (b) An example rollout on a real-world manipulation task, where actions selected by \algname with 3 iterations are directly executed at every step. \algname improves the robustness and precision of robot actions, enabling corrective behavior such as in Step 2. (c) An example rollout on RefCOCO questions.}
     \label{fig:rollouts}
\end{figure}

Our first set of real robot experiments evaluate \algname's ability to perform zero-shot robotic control with mobile manipulator navigation and manipulation, and Franka manipulation. 
These highlight the flexibility of \algname, as these robots vary in terms of control settings (navigation and manipulation), camera views (first and third person), as well as action space dimensionalities.
For example, Figure~\ref{fig:rollouts} illustrates several qualitative rollouts of \algname and the action samples (projected onto the images) as it steps through the iteration process. Note that after optimization, selected actions are more precisely positioned on target objects and areas of interest (most relevant to the input language instructions), without any model fine-tuning.
For goal-directed navigation tasks, we quantitatively evaluate \algname by measuring the success rates of whether it enables the mobile manipulator to reach its target destination (provided as a language input to \algname). For manipulation, we evaluate performance via three metrics (i) whether the robot end-effector reaches the relevant object (reach), (ii) efficiency via the number of action steps before successful termination (steps), and (iii) the success rate at which the robot grasps the relevant object (grasp -- when applicable).

\begin{table}[bht]
\footnotesize
\centering
\caption{\small Navigation success rate on the mobile manipulator in Figure~\ref{fig:embodiments} (a). We observe that iterations and parallel calls improve performance.}\label{table:nav_real}
\vspace{-0.5cm}
\begin{tabular}{c@{}cccc}\\
\toprule  
& No Iteration & 3 Iterations & No Iteration & 3 Iterations \\
Task & No Parallel & No Parallel & 3 Parallel & 3 Parallel \\
\midrule
Go to orange table with tissue box & 25\% & 50\% & \textbf{75\%} & \textbf{75\%}   \\ 
Go to wooden bench without hitting obstacle & 25\% & 50\% & \textbf{75\%} &  50\%   \\ 
Go to the darker room & 25\% & 50\% & 75\% & \textbf{100\%}  \\ 
Help me find a place to sit and write & 75\% & 50\% & \textbf{100\%} & 75\% \\ 
\bottomrule
\end{tabular}
\end{table}

Results on both navigation and manipulation tasks (shown in Tables~\ref{table:nav_real} and \ref{table:manip_real}) demonstrate that (i) \algname enables non-zero task success for both domains, (ii) parallel calls improves performance (in terms of success rates) and efficiency (by reducing the average number of actions steps), and (iii) increasing the number of \algname iterations also improves performance.
\begin{table}[hbt]
\footnotesize
\centering
\caption{\small Manipulation results on the mobile manipulator shown in Figure~\ref{fig:embodiments} (b), where ``Reach'' indicates the rate at which the robot successfully reached the relevant object, ``Steps'' indicates the number of steps, and ``Grasp'' indicates the rate at which the robot successfully grasped the relevant object (when applicable for the task).
We observe that while all approaches are able to achieve some non-zero success, iteration and parallel calls improve performance and efficiency of the policy. }\label{table:manip_real}
\vspace{-0.5cm}
\begin{tabular}{cccccccccc}\\
\toprule  
& \multicolumn{3}{c}{No Iterations} & \multicolumn{3}{c}{3 Iterations} & \multicolumn{3}{c}{3 Iterations} \\
 & \multicolumn{3}{c}{No Parallel} & \multicolumn{3}{c}{No Parallel} & \multicolumn{3}{c}{3 Parallel} \\
 \cmidrule(lr){2-4} \cmidrule(lr){5-7} \cmidrule(lr){8-10}
 Task & Reach & Steps & Grasp & Reach & Steps & Grasp & Reach & Steps & Grasp \\
\midrule
Pick coke can & 50\% & 4.5 & 0.0\% & 67\% & \textbf{3.0} & 33\% & \textbf{100\%} & \textbf{3.0} & \textbf{67\%} \\ 
Bring the orange to the X & 20\% & 4.0 & - & \textbf{80\%} & \textbf{3.5} & - & 67\% & \textbf{3.5} & - \\ 
Sort the apple & 67\% & 3.5 & - & \textbf{100\%} & 3.25 & - & 75\% & \textbf{3.0} & - \\ 
\bottomrule
\end{tabular}
\end{table}
Appendix~\ref{app:franka} and \ref{app:ravens} presents results on real Franka arm and a simulated RAVENS domain.

\subsection{Zero-shot Visual Grounding}\label{sec:exp_beyond}

In addition to robotic control tasks, we also examine \algname for reference localization tasks from RefCOCO~\cite{yu2016modeling}, which evaluates precise and robust visual grounding. 
To this end, we evaluate GPT-4V with 3 rounds of \algname on a random subset of 1000 examples from the RefCOCO testA split. We find strong performance even in the first iteration with modest improvement over further iterations. Prompts used are in Appendix \ref{app:prompts} and results are in Figure \ref{fig:refcoco} and examples in Figure~\ref{fig:rollouts}.

\begin{SCfigure}[][thb]
    \centering
    \begin{subfigure}[b]{0.27\textwidth}
    \centering
    \includegraphics[width=\linewidth]{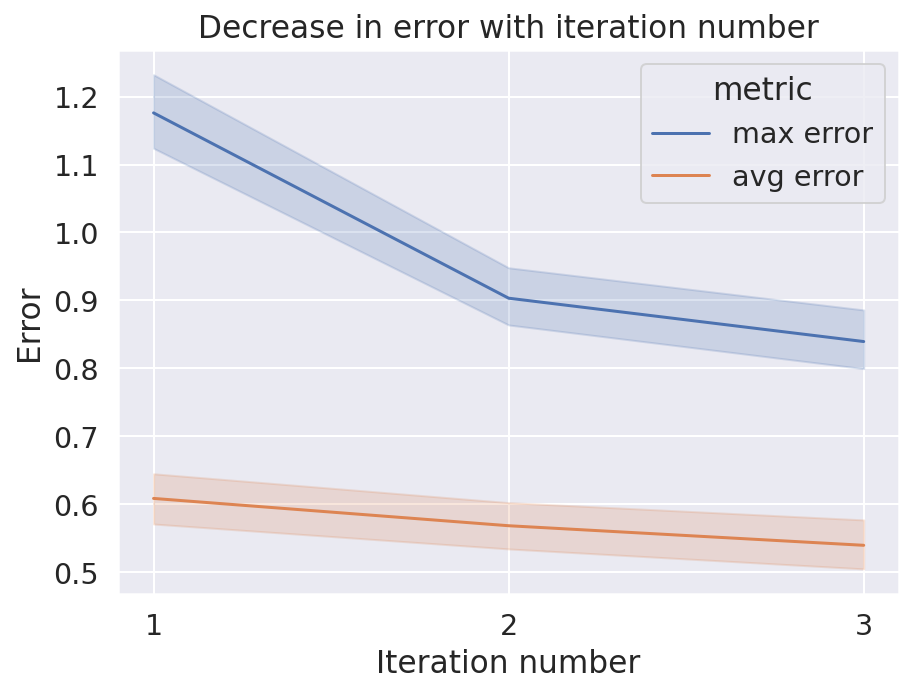}
    \end{subfigure}
    \begin{subfigure}[b]{0.27\textwidth}
    \centering
    \includegraphics[width=\linewidth]{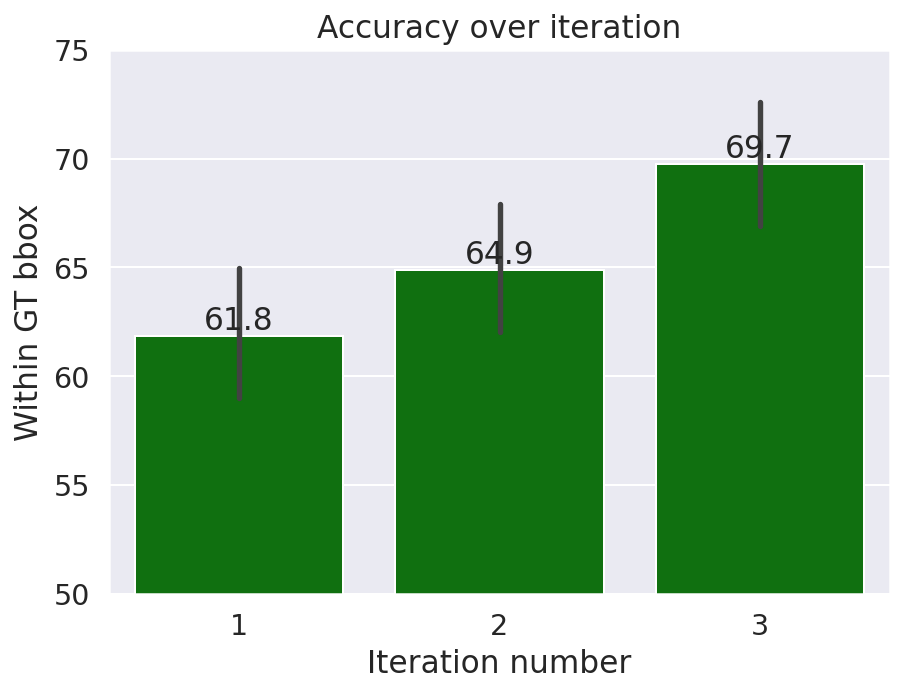}
    \end{subfigure}
    \caption{\small RefCOCO quantitative results. (Left) Normalized distance between the center of the ground truth bounding box and the selected circle. (Right) Accuracy as measured by whether the selected circle lies within the ground truth bounding box.}\label{fig:refcoco}
\end{SCfigure}

We provide an interactive demo on HuggingFace with a few demonstrative images as well as the ability to upload new images and questions; available \href{https://huggingface.co/spaces/pivot-iterative-visual-optimization/pivot-demo}{here}.

\subsection{Offline Performance and Ablations}\label{sec:exp_ablations}

In this section, we examine each element of \algname (the text prompt, visual prompt, and iterative optimization) through an offline evaluation, allowing a thorough evaluation without requiring execution on real robots.
To do this, we use demonstration data as a reference and compute how similar the action computed by \algname is to the ground-truth expert action.

For the manipulation domain, we obtain the reference robot action from the RT-X dataset~\cite{padalkar2023open} and compute the cosine similarity of the two actions in the camera frame as our metric. This metric measures how VLM choice is ``aligned" with human demonstrations. For example, a $0.5$ cosine similarity in 2D space correspond to $\arccos(0.5)= 60 ^{\circ}$.
As our actions can be executed a maximum delta along the chosen Cartesian action direction, we have found this metric more informative than others, e.g., mean squared error.
For the navigation domain, we use a human-labeled dataset from navigation logs and compute the normalized L2 distance between the selected action and the point of interest in camera frame as our metric. 
More information on each offline dataset can be found in Appendix~\ref{app:manip_eval} and \ref{app:nav_eval}.

\textbf{Text prompts.} To understand the effect of different text prompts, we experiment with several design choices, with numbers reported in Appendix~\ref{app:manip_eval}.
We investigate the role of zero-shot, few-shot, chain of thought, and direct prompting; we find that zero-shot chain of thought performs the best, though few-shot direct prompting is close and more token efficient.
We also experiment over the ordering of the image, preamble, and task; finding that preamble, followed by image, followed by task performs best, though by a small margin.

\begin{wrapfigure}{r}{0.5\textwidth}
    \centering
    \begin{subfigure}[b]{0.28\textwidth}
    \centering
    \includegraphics[width=\linewidth]{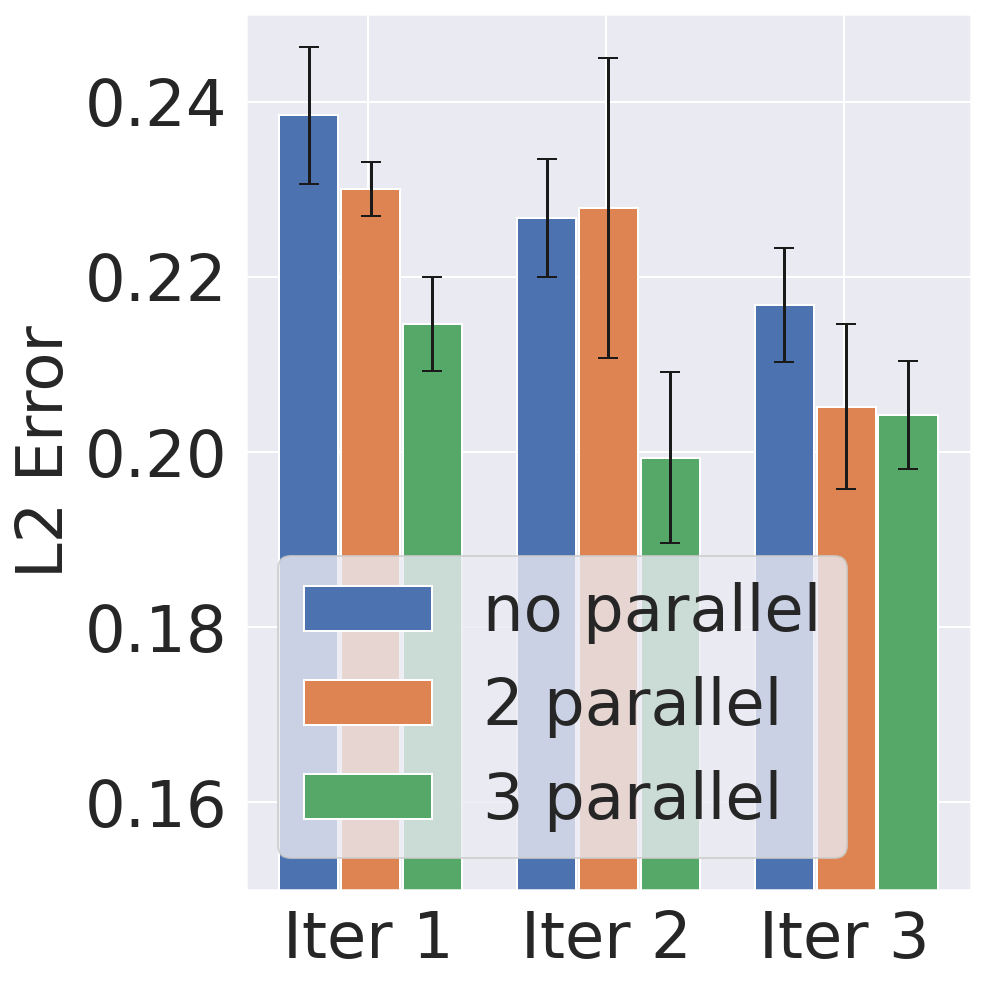}
    \caption{}\label{fig:nav-offline-barplot}
    \end{subfigure}
    \begin{subfigure}[b]{0.2\textwidth}
    \centering
    \includegraphics[width=\linewidth]{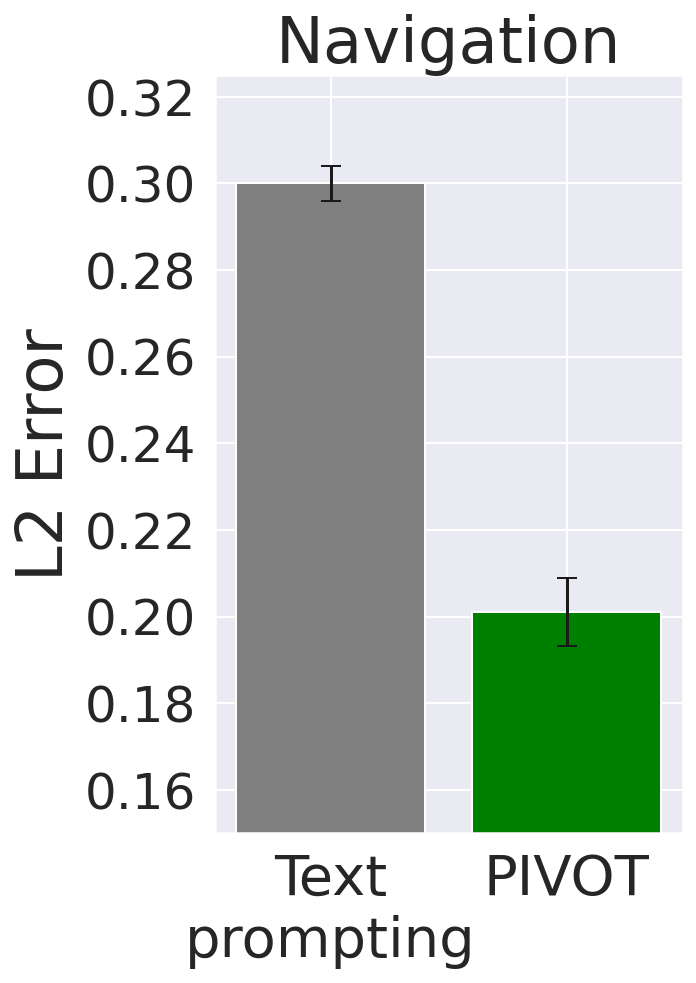}
    \caption{}\label{fig:ablation-language_nav}
    \end{subfigure}
    \caption{\small Offline evaluation results for navigation task with L2 distance (lower is better). 
    Ablation over (\ref{fig:nav-offline-barplot}) iterations and parallel calls and (\ref{fig:ablation-language_nav}) text-only baseline.}
    \label{fig:offline-navigation}
\end{wrapfigure}

\begin{figure}[htb]
    \centering
    \begin{subfigure}[b]{0.22\textwidth}
    \centering
    \includegraphics[width=\linewidth]{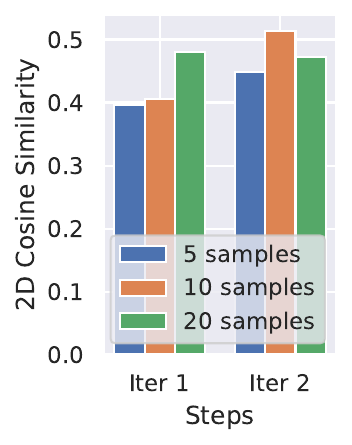}
    \caption{Number of samples}\label{fig:ablation-samples}
    \end{subfigure}
    \begin{subfigure}[b]{0.22\textwidth}
    \centering
    \includegraphics[width=\linewidth]{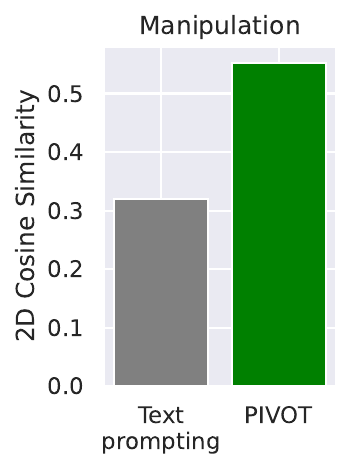}
    \caption{Text-only baseline}\label{fig:ablation-language}
    \end{subfigure}
    \begin{subfigure}[b]{0.22\textwidth}
    \centering
    \includegraphics[width=\linewidth]{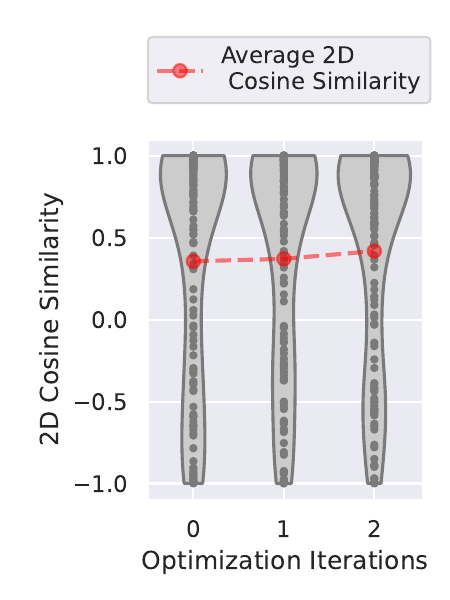}
    \caption{Iterations}\label{fig:ablation-iterations}
    \end{subfigure}
    \begin{subfigure}[b]{0.22\textwidth}
    \centering
    \includegraphics[width=\linewidth]{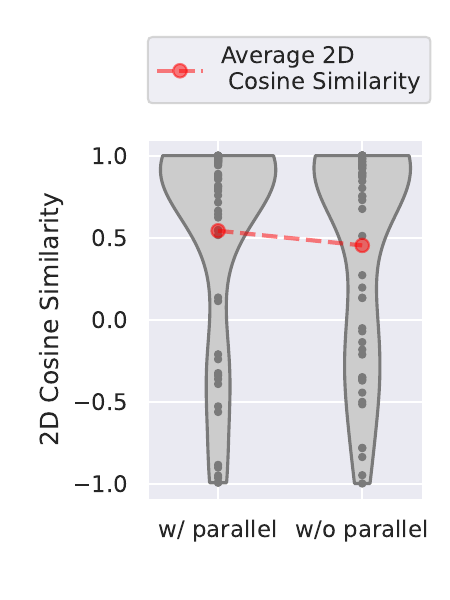}
    \caption{Parallel calls}\label{fig:ablation-parallel}
    \end{subfigure}
    \caption{\small Offline evaluation results for manipulation tasks with cosine similarity (higher is better). 
    }
    \label{fig:offline-manipulation}
\end{figure}

\textbf{Visual prompts.} 
Aspects of the style of visual prompts has been examined in prior works~\cite{yang2023set,shtedritski2023does}, such as the color, size, shading, and shape.
Herein, we investigate aspects central to \algname -- the number of samples and the importance of the visual prompt itself.
An ablation over the number of samples is shown in Figure~\ref{fig:offline-manipulation} where we note an interesting trend: more samples leads to better initial answers, but worse optimization.
Intuitively, a large number of samples supports good coverage for the initial answer, but with too many samples the region of the image around the correct answer gets crowded and causes significant issues with occlusions. 
For our tasks, we found 10 samples to best trade off between distributional coverage and maintaining sufficient visual clarity. 

To understand the necessity of the visual prompt itself, we compare to a language only baseline, where a VLM selects from a subset of language actions that map to robotic actions.
For the manipulation task, the VLM is given an image and task and selects from move ``right'', ```left'', ``up'', and ``down''.
A similar navigation benchmark is described in Appendix~\ref{app:nav_eval}.
We see in Figure~\ref{fig:offline-manipulation} and Figure~\ref{fig:offline-navigation} that \algname outperforms text by a large margin.
We note here that we do not compare to learned approaches that require training or finetuning as our focus is on zero-shot understanding. 
We believe many such approaches would perform well in distribution on these tasks, but would have limited generalization on out of distribution tasks.

\textbf{Iterative optimization.}
To understand the effect of the iterative optimization process, we ablate over the number of iterations and parallel calls.
In Figures~\ref{fig:refcoco}, \ref{fig:offline-navigation}, and \ref{fig:offline-manipulation}, we find that increasing iterations improves performance, increasing parallel calls improves performance, and crucially doing both together performs the best.
This echos the findings in the online evaluations above.

\subsection{Scaling}\label{sec:exp_scaling}

\begin{figure}[htb]
  \centering
  \includegraphics[width=0.27\textwidth]{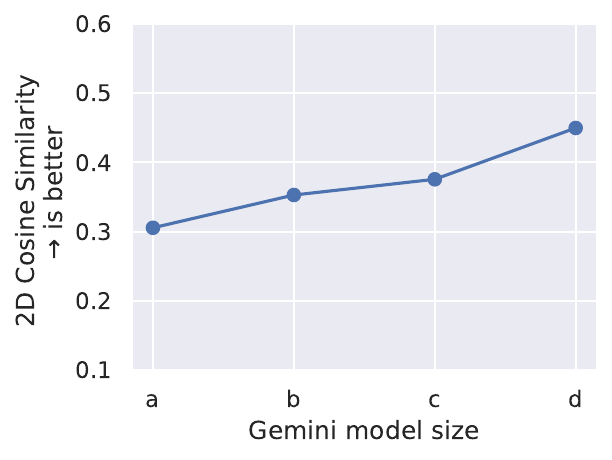}
  \includegraphics[width=0.27\textwidth]{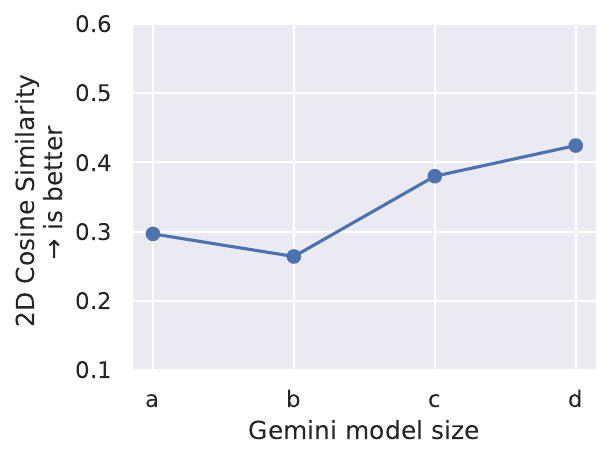}
  \includegraphics[width=0.28\textwidth]{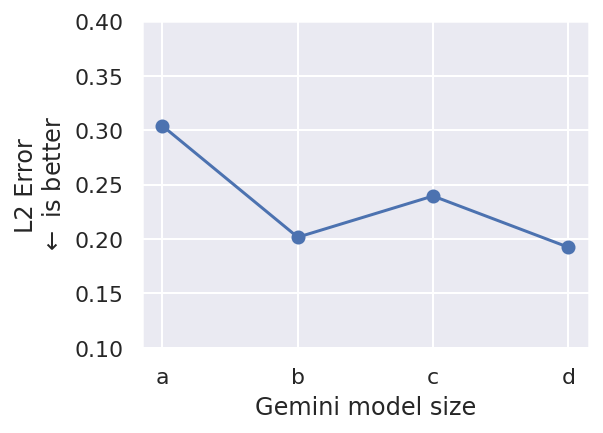}
  \caption{\small Scaling results of first iteration visual prompting performance across Gemini model \cite{gemini2023gemini} sizes show that \algname scales well with improved VLMs. Left and center plots are manipulation (pick up objects, move one object next to another), right plot is navigation.}
    \label{fig:scaling}
\end{figure}

We observe that \algname  scales across varying sizes of VLMs on the mobile manipulator offline evaluation (results measured in terms of cosine similarity and L2 error between \algname and demonstration data ground truth in Figure~\ref{fig:scaling}).
In particular, we compare \algname using four sizes of the Gemini family of models~\cite{gemini2023gemini} which we labeled a to d, with progressively more parameters. %
We find that performance increases monotonically across each model size.
Although there are still significant limitations and capabilities gaps, we see this scaling as a promising sign that \algname can leverage next-generation foundation models with increasing model size and capabilities \cite{gemini2023gemini}.

\subsection{Limitations}
\label{sec:exp_limitations}
In this work, we evaluate \algname using state-of-the-art VLMs and their zero-shot capabilities. 
We note that the base models have not been trained on in-domain data for robotic control or physical reasoning represented by visual annotation distributions.
While the exact failure modes may be specific to particular underlying VLMs, we continue to observe trends which may reflect broad limitation areas.
We expect that future VLMs with improved generalist visual reasoning capabilities will likewise improve in their visual annotation and robotics reasoning capabilities, and the general limitations of \algname on current state-of-the-art VLMs may serve to highlight potential risks and capabilities gaps, that point to interesting open areas for future work.

\textbf{3D understanding.}
While VLMs only take 2D images as visual inputs, in principle the image annotations and transformations applied via \algname can represent 3D queries as well.
Although we examined expressing depth values as part of the annotations using colors and label sizes (and described what they map to within a preamble prompt), we have observed that none of the VLMs we tested are capable of reliably choosing actions based on depth.
Beyond this, generalizing to higher dimensional spaces such as rotation poses even additional challenges.
We believe more complex visuals (e.g. with shading to give the illusion of depth) may address some of these challenges, but ultimately, the lack of 3D training data in the underlying VLM remains the bottleneck.
It is likely that training on either robot specific data or with depth images may alleviate these challenges.

\begin{figure}[htb]
\centering
 \begin{subfigure}[b]{0.46\textwidth}
    \centering
    \includegraphics[width=0.9\linewidth]{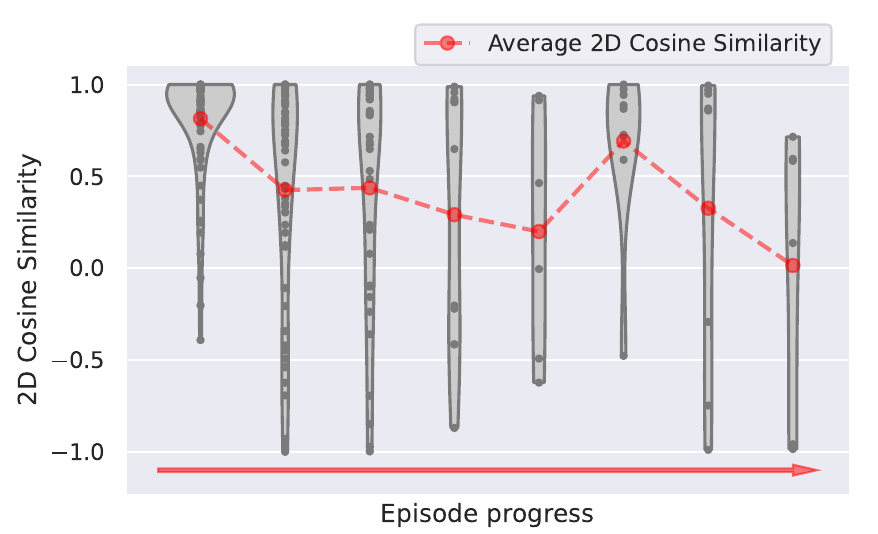}
\end{subfigure}
 \begin{subfigure}[b]{0.25\linewidth}
    \includegraphics[width=\linewidth]{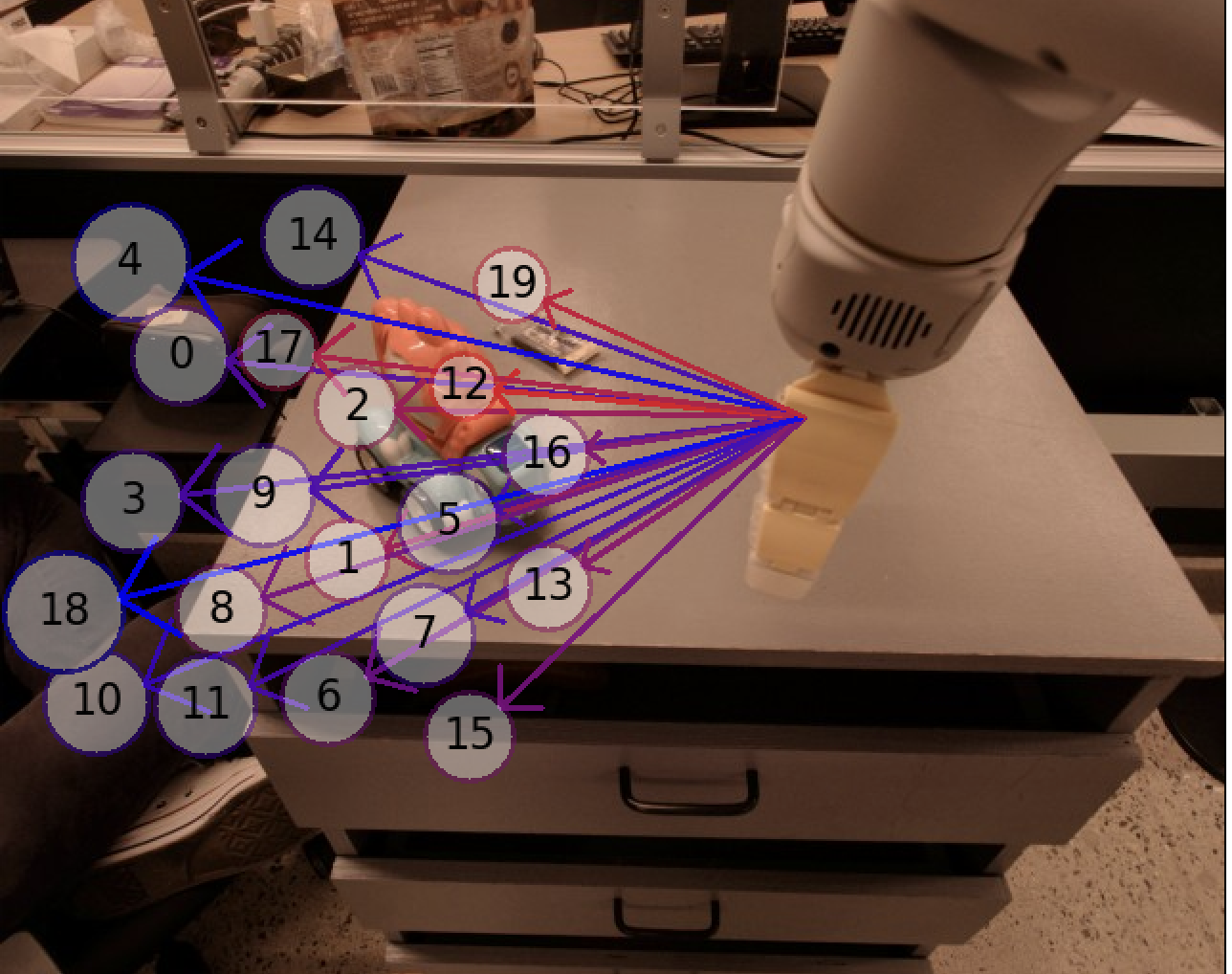}
    \caption{Easy scenario}%
\end{subfigure}
 \begin{subfigure}[b]{0.25\linewidth}
    \includegraphics[width=\linewidth]{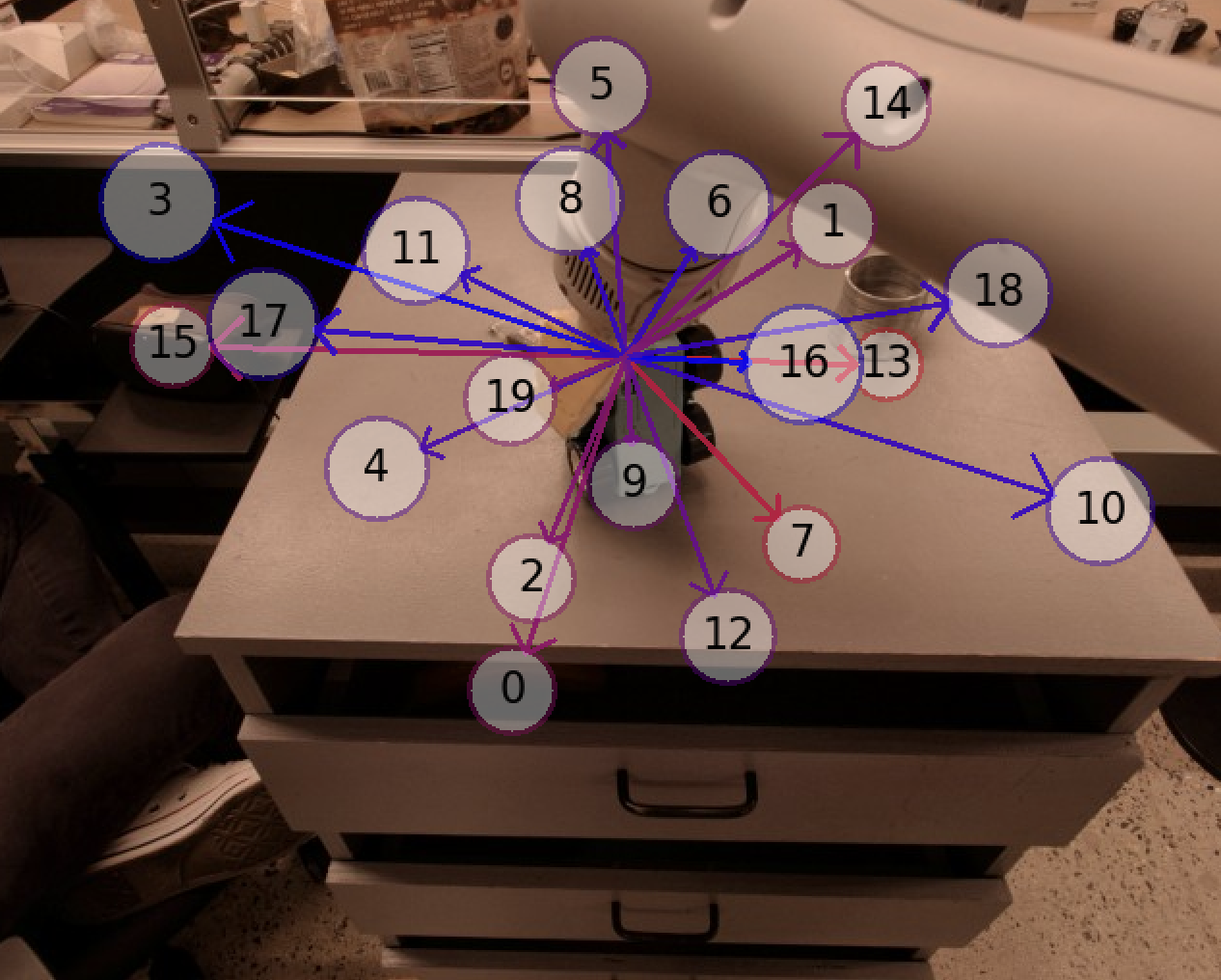}
    \caption{Hard scenario}%
\end{subfigure}
    \caption{\small \algname performance over ``move near'' trajectories, which pick up an object and move them near another. 
    Initially performance is high, but decreases as the robot approaches the grasp and lift (due to objects being obscured and the VLM not understanding the subtlety of grasping). After the grasp, the performance increases as it moves to the other object, but again decreases as it approaches.}
    \label{fig:interaction}
\end{figure}

\textbf{Interaction and fine-grained control.} 
During closed-loop visuomotor tasks (\eg for first-person navigation tasks, or manipulation task with hand-mounted cameras), images can often be characterized by increasing amounts of occlusion, where the objects of interest can become no longer visible if the cameras are too close. This affects \algname and the VLM's capacity for decision-making \eg determining when to grasp, whether to lift an object, or approaching an object from the correct side to push.
This is visualized in Figure~\ref{fig:interaction}, where errors over the trajectory are shown.
These errors are a result of both occlusions, resolution of the image, but perhaps more crucially, a lack of training data from similar interactions.
In this case, training on embodied or video data may be a remedy.

\textbf{Greedy behavior.}
Though we find iterative optimization alleviates many simple errors, we also find that the underlying VLM often displays greedy, myopic behaviors for multi-step decision-making tasks.
For instance, given the task ``move the apple to the banana'', the VLM may recommend immediately approaching the banana rather than the apple first.
We believe these mistakes may lessen with more capable VLMs, or with more in-domain examples provided either via fine-tuning or via few-shot prompting with \eg a history of actions as input context to the VLM to guide future generated acitons.

\textbf{Vision-language connection reasoning errors.}
We find that though overall the thought process of the VLM is reasonable, it stochastically connects the thought process to the incorrect arrow.
This issue appears to be a challenge of autoregressive decoding, once the number is decoded, the VLM must justify it, even if incorrect, and thus hallucinates an otherwise reasonable thought process.
Many of these errors are remedied through the optimization process of \algname, but we believe further improvements could be made with tools from robust optimization.

\section{Conclusion}

\algname presents a promising step towards leveraging VLMs %
for spatial reasoning zero-shot, and suggests new opportunities to cast traditionally challenging problems (e.g., low-level robotic control) as vision ones.
\algname can be used for tasks such as controlling a robot arm that require outputting spatially grounded continuous values with a VLM zero shot.
This is made possible by representing spatial concepts in the image space and then iteratively refining those by prompting a VLM.
Built on iterative optimization, \algname stands to benefit from other sampling initialization procedures, optimization algorithms, or search-based strategies.
Furthermore, we have identified several limitations of current state-of-the-art models that limits performance herein (\eg 3D understanding and interaction).
Therefore, adding datasets representing these areas presents an interesting avenue for future work; along with directly finetuning task specific data.
More importantly, though, we expect the capabilities of VLMs to improve over time, hence the zero-shot performance of \algname is likely to improve as well, as we have investigated in our scaling experiments.
We believe that this work can be seen as an attempt to unify internet-scale general vision-language tasks with physical problems in the real world by representing them in the same input space.
While the majority of our experiments focus on robotics, the algorithm can generally be applied to problems that require outputting continuous values with a VLM.

\subsection*{Acknowledgements}

We thank Kanishka Rao, Jie Tan, Carolina Parada, James Harrison, Nik Stewart, and Jonathan Tompson for helpful discussions and providing feedback on the paper.

\footnotesize
\bibliographystyle{icml2024}
\bibliography{cite}

\clearpage
\normalsize
\appendix
{\Large\textbf{Appendix}}
\input{appendix.tex}

\end{document}

%% file: appendix.tex
\section{Robotic Embodiments}\label{app:robots}

\textbf{Mobile Manipulator Navigation.} Shown in Figure~\ref{fig:embodiments} (a), we use a mobile manipulator platform for navigation tasks.
We use the image from a fixed head camera and annotate the image with arrows originating from the bottom center of the image to represent the 2D action space.
After \algname identifies the candidate action in the pixel space, we then use the on-board depth camera from the robot to map it to a 3D target location and command the robot to move toward the target (with a maximum distance of 1.0m).
We evaluate \algname on both a real robot and on an offline dataset. For real robot evaluation, we designed four scenarios where the robot is expected to reach a target location specified either through an object of interest (e.g.\ find apple) or through an indirect instruction (e.g.\ find a place to take a nap). For offline evaluation, we created a dataset of $60$ examples from prior robot navigation data with labeled ground truth targets. More details on the task and dataset can be found in Appendix Section~\ref{app:nav_eval}.

\textbf{Mobile Manipulator Manipulation.} Shown in Figure~\ref{fig:embodiments} (b), we use a mobile manipulator platform for manipulation tasks.
We use the image from a fixed head camera and annotate the image with arrows originating from the end-effector in camera frame, for which each arrow represents a 3D relative Cartesian end-effector position ($x, y, z$).
To handle the z-dimension height, we study two settings: one where height is represented through color grading (a red to blue spectrum) and one where the arm only uses fixed-height actions.
Gripper closing actions are not shown as visual annotations but instead expressed through text prompts.
Note that although the end-effector has rotational degrees of freedoms, we fix these due to the difficulty of expressing them with visual prompting, as is discussed in Section~\ref{sec:exp_limitations}.
We evaluate \algname on both real robot and an offline dataset.
For real robot evaluation, we study three tabletop manipulation tasks which require combining semantic and motion reasoning.
Success criteria consists of binary object reaching success, number of steps taken for successful reaching trajectories, and grasping success when applicable.
For offline evaluation, we use demonstration data from the RT-X mobile manipulator dataset~\cite{padalkar2023open}.
We sample 10 episodes of pick demonstrations for most of our offline evaluations, and 30 episodes of move near demonstrations for our interaction Figure~\ref{fig:interaction}.
More details on the results can be found in Appendix Section~\ref{app:manip_eval}.

\textbf{Franka.} Shown in Figure~\ref{fig:embodiments} (c) we use the Franka for manipulation.
We use the image from a wrist mounted camera and annotate the image with arrows originating from the center of the camera frame, for which each arrow represents a 3D relative Cartesian end-effector position ($x, y, z$, where the $z$ dimension is captured with a color spectrum from red to blue). 
We examine both pick tasks and place tasks, with 5 objects for each task. 
More details on the results can be found in Appendix Section~\ref{app:franka}.

\textbf{RAVENS~\cite{zeng2021transporter}.} Show in Figure~\ref{fig:embodiments} (d), we use the RAVENS simulation domain for pick and place manipulation.
We use the image from an overhead camera and annotate the image with pick and place locations, following the action representation in \citet{zeng2021transporter}.
This action space allows us to evaluate higher-level action representations.
More details on the results can be found in Appendix Section~\ref{app:ravens}.

\section{Mobile Manipulator Navigation Offline Evaluation}\label{app:nav_eval}

\subsection{Dataset}

\begin{figure}[h]
    \centering
    \includegraphics[width=\linewidth]{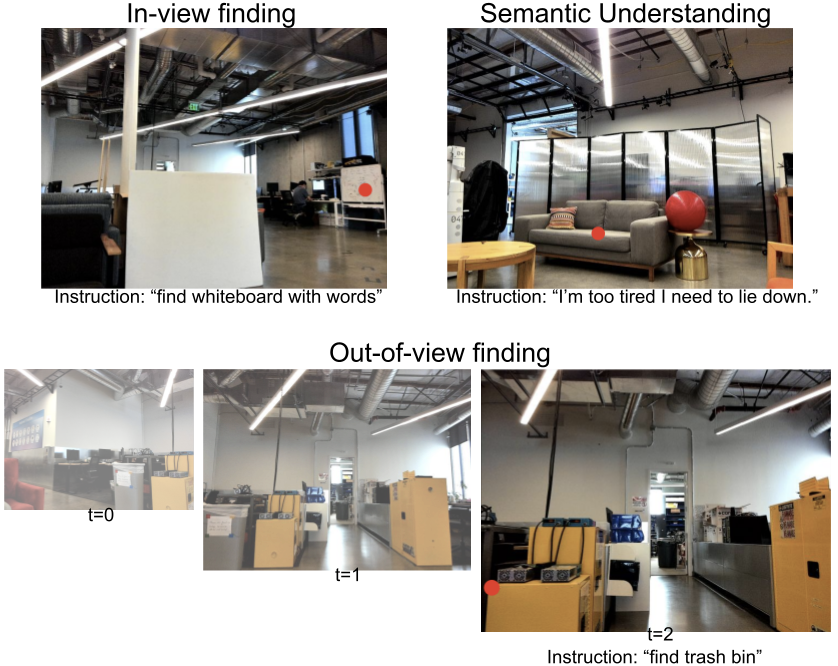}
    \caption{\small Example tasks in the offline navigation dataset from different task categories. Red dot denotes the ground truth target.}
    \label{fig:nav-offline-tasks}
\end{figure}

We create an offline dataset of $60$ examples using images collected from the on-robot camera sensor by walking the robot in an indoor environment. For each example, we provide an instruction and a associated location in the image space as the target. We categorize our tasks into three types: 1) in-view finding, where the robot is tasked to approach an object within the line of sight, 2) semantic understanding, where the instruction implicitly refers to an object in view 3) out-of-view finding, where the object of interest is not visible from the current view with arrow annotations, but can be seen in past images from different locations. Figure \ref{fig:nav-offline-tasks} shows examples of the three task categories. 

\subsection{Evaluation Results}

Table \ref{tab:nav_offline} shows the detailed evaluation results of \algname on the offline navigation dataset. We measure the accuracy of the \algname output by its deviation from the target point in image space normalized by the image width and break it down into the three task categories. We report mean and standard deviation for three runs over the entire dataset.

As seen in the table, by using the parallel call to robustify the VLM output we see significant improvements over running VLM only once (0 parallel) and running \algname for multiple iterations also improves accuracy of the task. However, increasing the parallel calls or the iteration number further did not achieve notably better performance.

\begin{table}[h]
\centering
\small
\caption{\small Navigation offline evaluation  measured in L2 loss (lower the better).}\label{tab:nav_offline}
\begin{tabular}{cccc}\\\toprule \multicolumn{4}{c}{In-View Tasks} \\\midrule
 & 1 iter  & 2 iter  & 3 iter \\\midrule
0 parallel & $0.21\pm0.002$ & $0.21\pm0.007$ & $0.19\pm0.007$ \\ 2 parallel & $0.19\pm0.004$ & $0.2\pm0.012$ & $0.18\pm0.005$\\
3 parallel & $0.19\pm0.003$ &$0.17\pm0.007$ & $0.17\pm0.009$\\ \bottomrule
\multicolumn{4}{c}{Semantic Tasks} \\\midrule
 & 1 iter  & 2 iter  & 3 iter \\\midrule
0 parallel & $0.23\pm0.012$ & $0.2\pm0.006$ & $0.19\pm0.025$ \\ 
2 parallel & $0.26\pm0.015$ & $0.21\pm0.02$ & $0.2\pm0.02$ \\ 
3 parallel & $0.21\pm0.01$ &$0.19\pm0.04$ & $0.19\pm0.01$ \\  \bottomrule
\multicolumn{4}{c}{Out-of-View Tasks} \\\midrule
 & 1 iter  & 2 iter  & 3 iter \\\midrule
0 parallel & $0.44\pm0.04$ & $0.38\pm0.015$ & $0.39\pm0.032$ \\
2 parallel & $0.38\pm0.001$ & $0.39\pm0.02$ & $0.39\pm0.02$ \\
3 parallel & $0.37\pm0.01$ &$0.38\pm0.026$ & $0.39\pm0.05$ \\  \bottomrule
\end{tabular}
\end{table} 

We compared our proposed approach, which reasons in image-space with image annotations, with reasoning in text without annotated images.
In this text-based baseline, we provide the same image and navigation query to the VLM, but we ask the VLM to imagine that the image is split into 3 rows and 3 columns of equal-sized regions and output the name of one of those regions (e.g. ``top left", "bottom middle").
We then compute the distance between the center of the selected region to the ground truth target point.
Given that we are not performing iterative optimization with this text baseline, we compare its results against \algname with just 1 iteration and 0 parallel.
See results in Table~\ref{tab:pivot_vs_text}.
For GPT-4V, the text baseline incurs higher mean and standard deviation of errors across all tasks.

\begin{table}[h]
    \centering
    \small
    \caption{\small Reasoning with Image Annotations vs. with Text for Navigation offline evaluations measured in L2 loss (lower the better).}
    \begin{tabular}{lccc}
    \toprule
    Method & In-View & Semantic & Out-of-View \\
    \midrule
    Image & $0.21\pm0.002$ & $0.23\pm0.012$ & $0.44\pm0.04$ \\
    Text & $0.26\pm0.15$ & $0.35\pm0.14$ & $0.46\pm0.31$ \\
    \bottomrule
    \end{tabular}
    \label{tab:pivot_vs_text}
\end{table}

\section{Mobile Manipulator Manipulation Online Evaluation}~\label{app:manip_online_eval}
\begin{figure*}[!t]
    \centering
    \includegraphics[width=\linewidth]{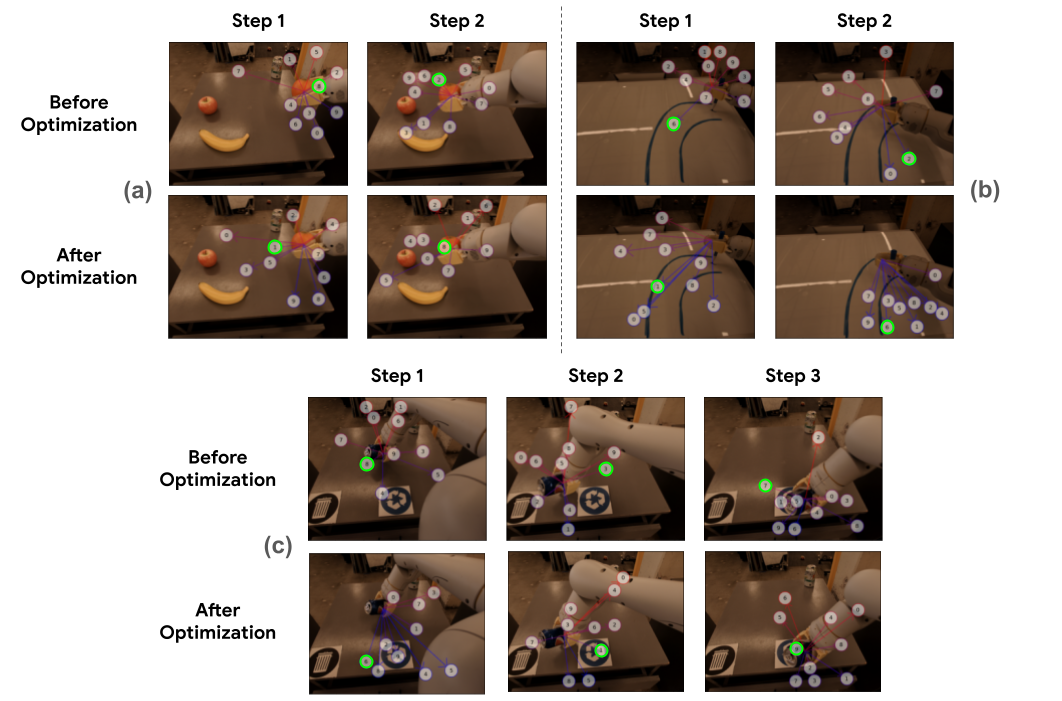}
    \caption{\small Evaluating \algname on real world mobile manipulator tabletop manipulation scenarios which require a combination of semantic reasoning and action understanding. Using 3 optimization iterations on the real world mobile manipulator, we see promising successes for (a) ``move the orange to complete the smiley face represented by fruits'', (b) ``use the marker to trace a line down the blue road'', and (c) ``sort the object it is holding to the correct piece of paper''.}
    \label{fig:manip-emergent}
\end{figure*}

In addition to the quantitative evaluation trials for the real-world manipulation experts described in Section~\ref{sec:exp_on_robot}, we also showcase additional evaluation rollouts in Figure~\ref{fig:manip-emergent}. Qualitatively, we find that \algname is able to recover from inaccuracies in action prediction, such as those which may result from imperfect depth perception or action precision challenges.

\section{Mobile Manipulator Manipulation Offline Evaluation}~\label{app:manip_eval}

\begin{figure}[htb]
    \centering
    \includegraphics[width=.45\linewidth]{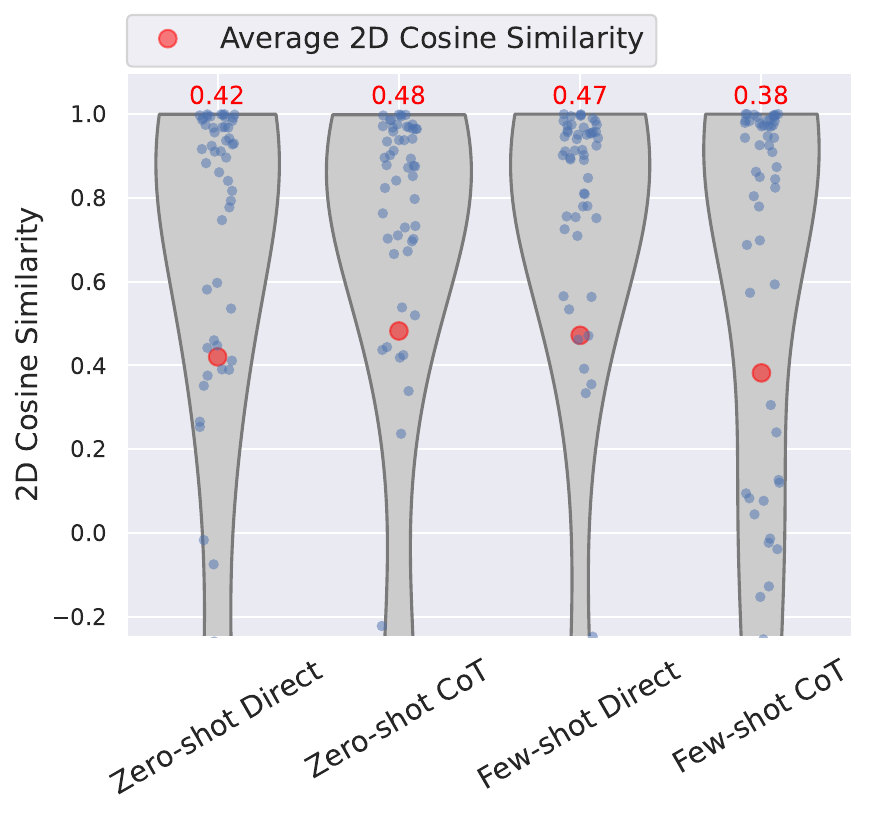}
    \includegraphics[width=.45\linewidth]{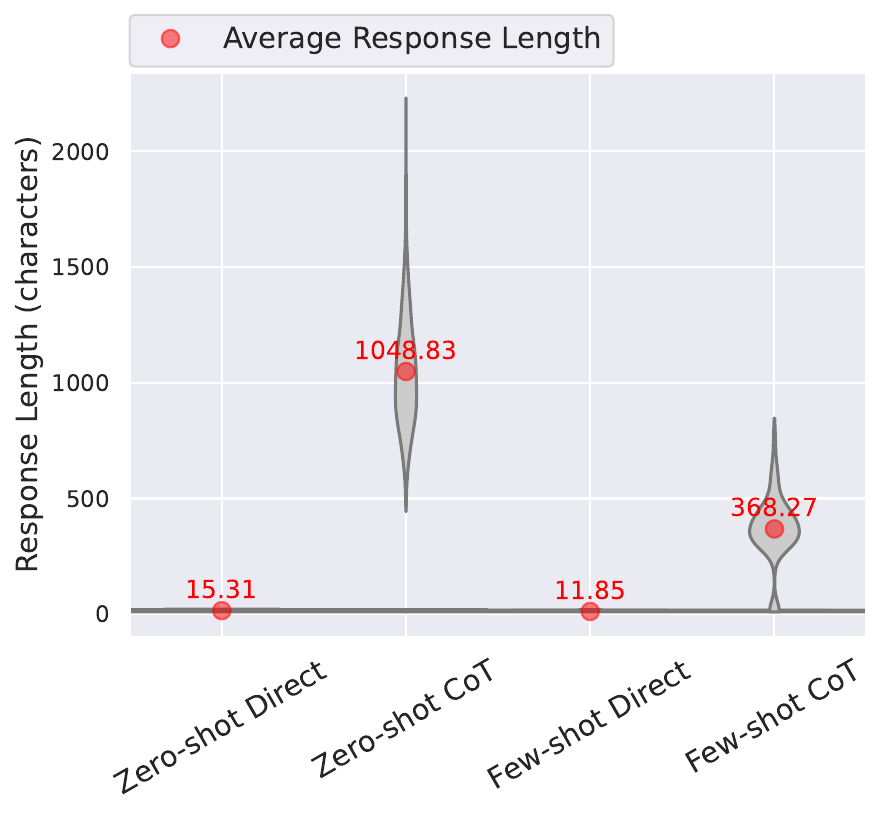}
    \caption{\small Ablation of few-shot vs.\ zero-shot and CoT vs. direct performance on manipulation domain. The best performing combination is zero-shot CoT. However, direct models can achieve similar performance with much fewer output tokens thus more token efficient.}
    \label{fig:ablation-text}
\end{figure}

\begin{figure}[htb]
    \centering
    \includegraphics[width=.45\linewidth]{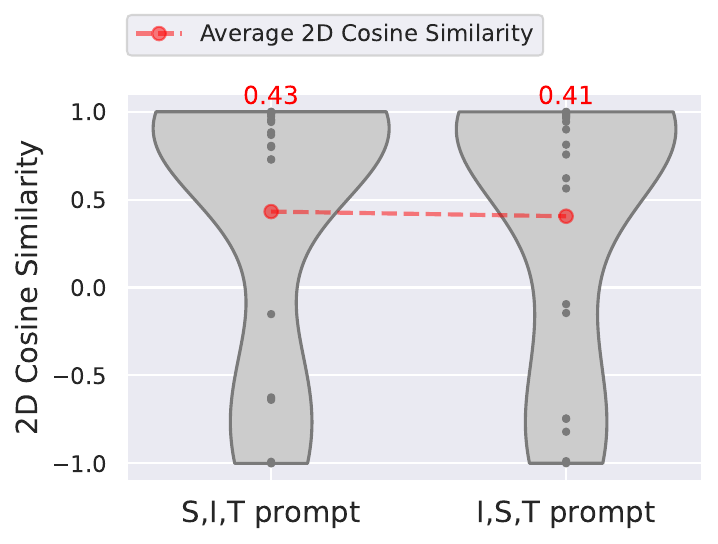}
    \caption{\small Ablation of order of preamble, image, and task on mobile manipulation domain. We found it is beneficial to put the image closer to the end of the prompt, though the effect is marginal. P, I, T means preamble, followed by image and task description, and I, P, T means image followed by preamble and task description.}
    \label{fig:ablation-order}
\end{figure}

Using the offline mobile manipulator dataset described in Section~\ref{app:robots}, we additionally ablate the text prompt herein. 
In Figure~\ref{fig:ablation-text} we consider the performance of zero-shot and few-shot prompting as well as chain of thought~\cite{wei2022chain} and direct prompting.
We find in general that neither is a panacea, though zero-shot chain of thought performs best, few-shot direct prompting performs similarly and is significantly more token efficient.
In Figure~\ref{fig:ablation-order} we consider the effect that the order of the prompt has on performance. 
The distinct elements of the prompt are the preamble (which describes the high level goal), the task (which describes the specific task the robot is attempting to perform), and the image.
Examples of these prompts can be seen in Appendix Section~\ref{app:prompts}.
We find a small amount of variation in performance between orders, with preamble, image, and task resulting in the highest performance.
We hypothesize that this order most closely mirrors the training mixture.

\begin{figure}[H]
    \centering
    \includegraphics[width=.95\linewidth]{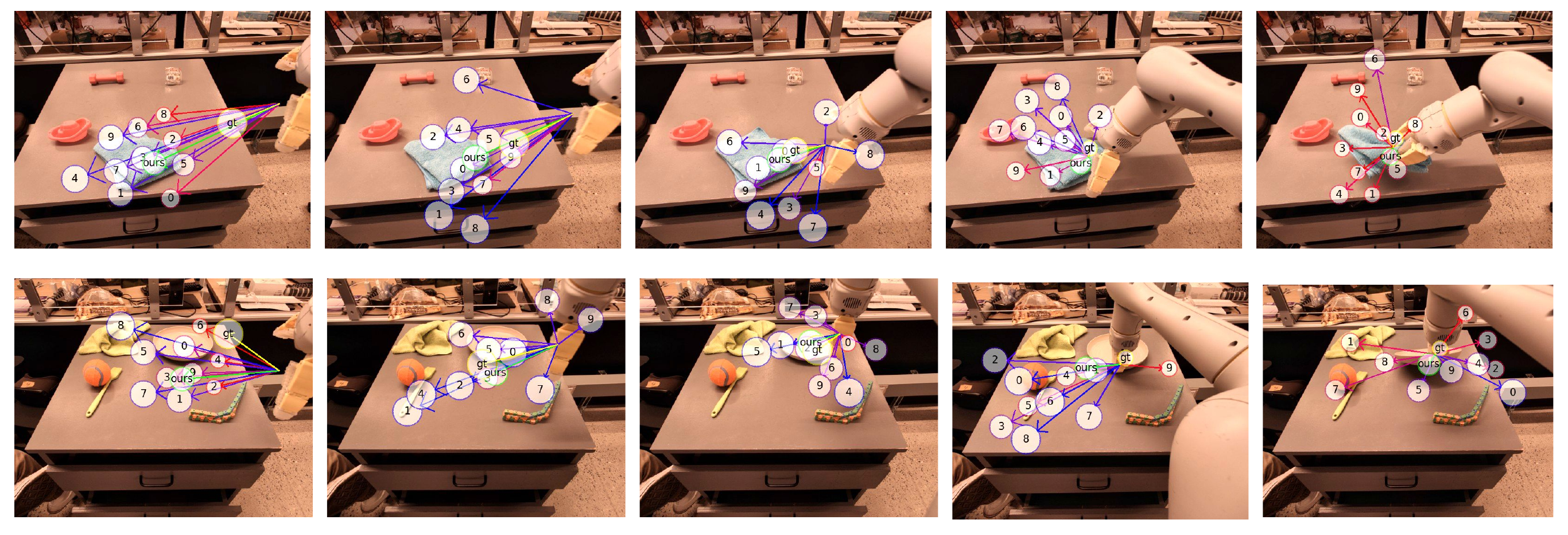}
    \caption{\small Two episodes of mobile manipulator manipulation offline evaluation. It shows our method can generate reasonable actions following the arrow annotations.}
    \label{fig:pivot-meta-offline}
\end{figure}

To illustate the limitation of our method decribed in Fig.~\ref{fig:interaction} better, we visualize two episodes of the mobile manipulator manipulation offline eval in Fig.~\ref{fig:pivot-meta-offline}. The figure shows that at the beginning of the episode where it is clear where to move, our method tend to generate accurate predictions while in the middle of the episode where there are interactions, our method struggles to generate correct actions.

\begin{figure}[htb]
    \centering
    \includegraphics[width=.95\linewidth]{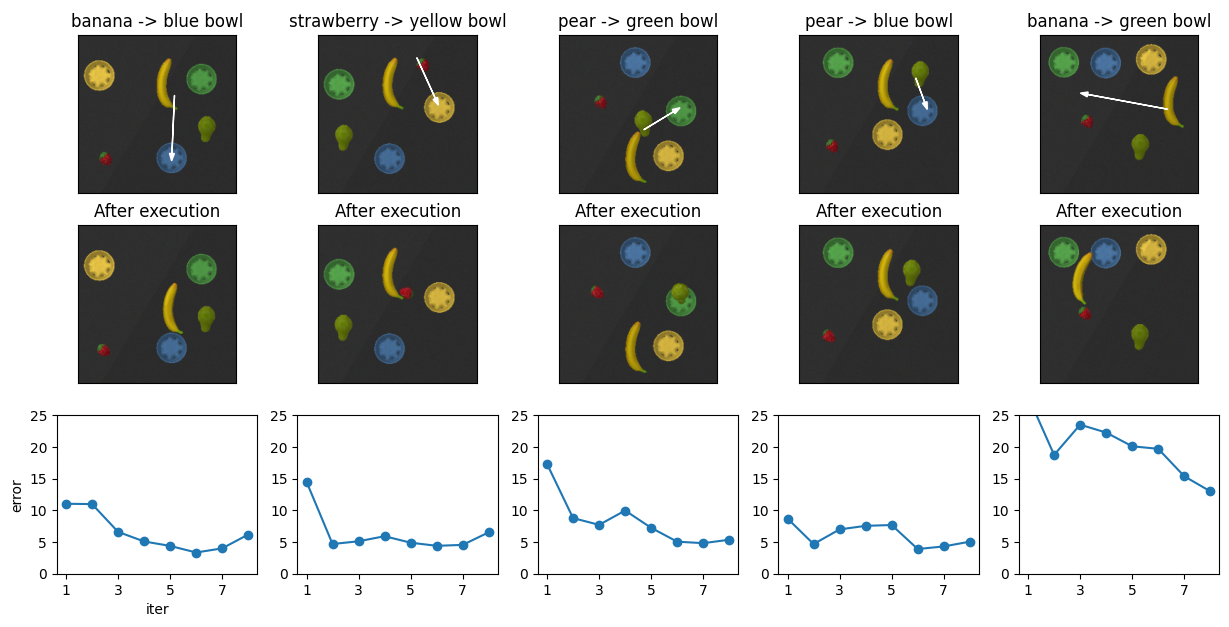}
    \caption{\small RAVENS evaluations. Each column shows a different task instance. Title: pick object followed by place object. Top row: initial image with pick and place locations predicted by VLM indicated by white arrow. Middle row: result after executing action. Bottom row: L2 distance between predicted and ground truth locations (averaged for both pick location and place location), over iterations.}
    \label{fig:ravens-evals}
\end{figure}

\section{RAVENS Online Simulation Evaluation}~\label{app:ravens}
We create a suite of evaluation tasks in which the robot must pick a specified fruit and place it in a specified bowl. There are three fruits in the scene (banana, strawberry, pear) and three bowls with different colors (blue, green, yellow).
Each task takes the form "pick the \{fruit\} and place it in the \{color\} bowl."
Given the task goal, we parse the source object and the target object, and independently prompt the VLM to get the pick and place locations corresponding to these two objects respectively.
Refer to Appendix~\ref{app:prompts} for the prompt we use.
In Figure~\ref{fig:ravens-evals} we report evaluation over five random instances.
Here we specifically report the error with respect to ground truth pick and place locations over each iteration of visual prompting.
We see that the error generally decreases in the first few iterations and eventually converges.
In most settings the chosen pick and place locations are close to the desired objects, yet the VLM lacks the often ability to precisely choose points that allow it to execute the task successfully in one action.

\section{Franka Online Evaluation}~\label{app:franka}
\begin{figure}[htb]
    \centering
    \includegraphics[width=.45\linewidth]{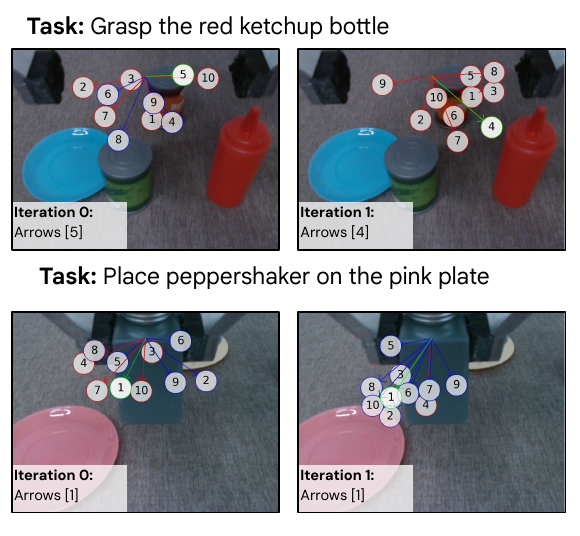}
    \caption{\small Rollouts on the Franka environment.}
    \label{fig:franka-evals}
\end{figure}

We evaluate \algname in a real world manipulation setting using a Franka robot arm with a wrist-mounted camera and a 4D relative Cartesian delta action space.
We study 7 tabletop manipulation tasks involving grasping and placing various objects, and analyze three version of \algname with varying numbers of optimization iterations and number of parallel \algname processes.
Each task is evaluated for two trials, for which we record intermediate reaching success rates for reaching the correct XY and YZ proximities for the target object (where in the camera frame the $x$-axis is into and out of the page, the $y$-axis is left and right, and the $z$ axis is up and down), as well as the overall number of timesteps taken for successful trials.
As shown in Table~\ref{table:franka_real}, we find that all instantiations of \algname are able to achieve non-zero success, but increasing the number of optimization iterations and number of parallel processes increases performance and stability.
Rollouts are shown in Figure~\ref{fig:franka-evals}.

\begin{table}[htb]
\footnotesize
\centering
\caption{\small Manipulation results on the real-world Franka setting shown in Figure~\ref{fig:embodiments} (c), where ``XY'' and ``YZ'' indicate success rates for reaching the relevant object XY and YZ proximities respectively and ``Steps'' indicates the number of steps taken if successful finished the task.
We observe that while all approaches are able to achieve some non-zero success, iteration and parallel calls improve performance and efficiency of the policy.
}\label{table:franka_real}
\vspace{-0.5cm}
\begin{tabular}{cccccccccc}\\
\toprule  
& \multicolumn{3}{c}{No Iterations} & \multicolumn{3}{c}{3 Iterations} & \multicolumn{3}{c}{3 Iterations} \\
 & \multicolumn{3}{c}{No Parallel} & \multicolumn{3}{c}{No Parallel} & \multicolumn{3}{c}{3 Parallel} \\
 \cmidrule(lr){2-4} \cmidrule(lr){5-7} \cmidrule(lr){8-10}
 Task & XY & YZ & Steps & XY & YZ & Steps & XY & YZ & Steps \\
\midrule
Place saltshaker on the blue plate & 0\% & 0\% & - & 0.5\% & 0\% & - & 50\% & 50\% & 3.0 \\
Place peppershaker on the pink plate & 100\% & 100\% & 8.0 & 100\% & 100\% & 3.5 & 50\% & 50\% & 4.0 \\
Grasp the pink cup & 50\% & 50\% & 7.0 & 0\% & 50\% & - & 0\% & 50\% & - \\
Grasp the pepper shaker & 50\% & 50\% & 8.0 & 0\% & 50\% & - & 0\% & 50\% & - \\
Grasp the blue cup & 0\% & 50\% & - & 0\% & 50\% & - & 0\% & 50\% & - \\
Grasp the red ketchup bottle & 0\% & 50\% & - & 0\% & 0\% & - & 100\% & 100\% & 6.0 \\
Grasp the can & 0\% & 0\% & - & 0\% & 0\% & - & 50\% & 50\% & 3.0 \\
\midrule
\textbf{Average} & 25\% & 38\% & 7.8 & 28\% & 31\% & 3.5 & 34\% & 59\% & 4.4 \\
\bottomrule
\end{tabular}

\end{table}

\section{Visual Annotation Sensitivity}
\label{app:arrows}
\begin{figure}
    \centering
    \includegraphics[width=0.55\linewidth]{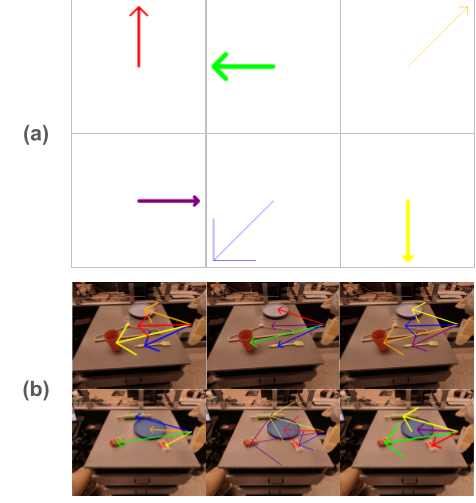}
    \caption{Examples of procedurally generated datasets studying the robustness of VLMs for understanding visual annotation arrow styles. (a) focuses on absolute direction understanding of single arrows on blank backgrounds. (b) focuses on object-relative arrow understanding in realistic scenes.}
    \label{fig:arrow_styles}
\end{figure}

Inspired by prior works which find interesting biases and limitations of modern VLMs on understanding visual annotations~\citep{shtedritski2023does,yang2023set,yang2023dawn}, we analyze the ability of state-of-the-art VLMs to understand various types of arrow annotations.
We generate two synthetic datasets: one toy dataset of various styles of CV2~\citep{itseez2015opencv} arrows overlaid on a white background, and a more realistic dataset of various styles of object-referential arrows overlaid on a real-world robotics scene.
The datasets adjust parameters such as arrow color, arrow thickness, and relative arrowhead size.
In the first dataset, we query VLMs to classify the direction of the arrows, which studies the effect of styling on the ability of VLMs to understand absolute arrow directions; examples are shown in Figure~\ref{fig:arrow_styles}.
In the second dataset, we query VLMs to select the arrow which points at a specified object out of multiple objects, which studies the effect of styling on the ability of VLMs to understand relative and object-centric arrow directions.
The second dataset contains scenes with various objects, which we categorize into ``Easy'' (plates, boxes, cubes), ``Medium'' (cups, bags, mugs), ``Hard'' (hangers, toys), and ``Very Hard'' (brushes, eccentric objects). 

\begin{table*}[t]
\footnotesize
\centering
\caption{\small Visual annotation arrow robustness of VLMs on a synthetic toy arrow dataset. For various colored arrows with different thicknesses, different sized arrowheads, and different absolute directions, we evaluate the robustness of GPT-4V on correctly classifying the absolute arrow direction.}\label{table:arrow_ablation1}
\vspace{-0.5cm}
\begin{tabular}{cccc|ccc|cccc}\\
\toprule  
& \multicolumn{3}{c}{Arrow Thickness} & \multicolumn{3}{c}{Arrowhead Size} & \multicolumn{4}{c}{Direction} \\
 \cmidrule(lr){2-4} \cmidrule(lr){5-7} \cmidrule(lr){8-11}
 Color & 2 & 4 & 6 & 0.1 & 0.3 & 0.5 & up+right & down+right & up+left & down+left \\
\midrule
red & 96\% & 92\% & 96\% & 97\% & 94\% & 88\% & 100\% & 75\% & 75\% & 92\% \\
orange & 92\% & 88\% & 96\% & 100\% & 91\% & 84\% & 100\% & 100\% & 50\% & 83\% \\
yellow & 88\% & 88\% & 100\% & 100\% & 94\% & 84\% & 93\% & 100\% & 75\% & 67\% \\
green & 96\% & 92\% & 96\% & 100\% & 100\% & 88\% & 100\% & 92\% & 92\% & 83\% \\
blue & 92\% & 92\% & 88\% & 91\% & 91\% & 88\% & 100\% & 17\% & 100\% & 100\% \\
purple & 100\% & 96\% & 96\% & 97\% & 97\% & 97\% & 100\% & 92\% & 92\% & 92\% \\
\bottomrule
\end{tabular}
\end{table*}

\begin{table*}[t]
\footnotesize
\centering
\caption{\small Visual annotation arrow robustness of VLMs on an object-referential arrow dataset. For various colored arrows with different thicknesses, different sized arrowheads, and different absolute directions, we evaluate the robustness of GPT-4V on correctly selecting the arrow which refers to a specified object.}\label{table:arrow_ablation2}
\vspace{-0.5cm}
\begin{tabular}{cccc|ccc|cccc}\\
\toprule  
& \multicolumn{3}{c}{Arrow Thickness} & \multicolumn{3}{c}{Arrowhead Size} & \multicolumn{4}{c}{Target Object} \\
 \cmidrule(lr){2-4} \cmidrule(lr){5-7} \cmidrule(lr){8-11}
 Color & 2 & 4 & 6 & 0.1 & 0.3 & 0.5 & Easy & Medium & Hard & Very Hard \\
\midrule
red & 42\% & 33\% & 33\% & 50\% & 33\% & 25\% & 44\% & 100\% & 0\% & 0\% \\
orange & 25\% & 25\% & 25\% & 25\% & 25\% & 25\% & 0\% & 100\% & 0\% & 0\% \\
yellow & 67\% & 58\% & 50\% & 83\% & 58\% & 33\% & 100\% & 33\% & 56\% & 44\% \\
green & 50\% & 58\% & 50\% & 83\% & 58\% & 33\% & 100\% & 33\% & 56\% & 44\% \\
blue & 42\% & 36\% & 33\% & 36\% & 50\% & 25\% & 100\% & 33\% & 22\% & 0\% \\
purple & 33\% & 50\% & 50\% & 58\% & 58\% & 17\% & 89\% & 22\% & 56\% & 11\% \\
\bottomrule
\end{tabular}
\end{table*}

\clearpage
\clearpage

\section{Prompts}
\label{app:prompts}
\subsection{RefCOCO prompt}
\noindent\fbox{\parbox{0.97\linewidth}{\footnotesize{\texttt{{%
Your goal is to find the OBJECT in this scene. I have annotated the image with numbered circles. Choose the 3 numbers that have the most overlap with the OBJECT. If there are no points with overlap, then don't choose any points. You are a five-time world champion in this game. Give a one sentence analysis of why you chose those points. Provide your answer at the end in a json file of this format:\\
\{"points": []
  \}
}
}}}}\\

\subsection{Navigation prompt}
\noindent\fbox{\parbox{0.97\linewidth}{\footnotesize{\texttt{{%
I am a wheeled robot that cannot go over objects. This is the image I'm seeing right now. I have annotated it with numbered circles. Each number represent a general direction I can follow. Now you are a five-time world-champion navigation agent and your task is to tell me which circle I should pick for the task of: \{INSTRUCTION\}? Choose \{K\} best candidate numbers. Do NOT choose routes that goes through objects.  Skip analysis and provide your answer at the end in a json file of this form:
\{"points": []
  \}
  }
}}}}\\

\subsection{RAVENS prompt}
\noindent\fbox{\parbox{0.97\linewidth}{\footnotesize{\texttt{{%
which number markers are closest to the \{OBJECT\}? Reason and express the final answer as 'final answer` followed by a list of the closest marker numbers.
  }
}}}}\\

\subsection{Manipulation online eval prompt}
\textbf{Direct}

\vspace{0.3em}
\noindent\fbox{\parbox{0.97\linewidth}{\footnotesize{\texttt{{
What number arrow should the robot follow to {task}?
}}}}}\\

\vspace{0.3em}
\noindent\fbox{\parbox{0.97\linewidth}{\footnotesize{\texttt{{
Rules:
- You are looking at an image of a robot in front of a desk trying to arrange objects. The robot has an arm and a gripper with yellow fingers.
- The arrows in the image represent actions the robot can take.
- Red arrows move the arm farther away from the camera, blue arrows move the arm closer towards the camera.
- Smaller circles are further from the camera and thus move the arm farther, larger circles are closer and thus move the arm backwards.
- The robot can only grasp or move objects if the robot gripper is close to the object and the gripper fingers would stably enclose the object
- Your answer must end with a list of candidate arrows which represent the immediate next action to take (~0.3 seconds). Do not consider future actions between the immediate next step.
- If multiple arrows represent good immediate actions to take, return all candidates ranked from worst to best.
- A general rule of thumb is to return 1-4 candidates.
Instruction: Reason through the task first and at the end summarize the correct action choice(s) with the format, ``Arrow: [<number>, <number>, etc.].``
Task: {task}
}}}}}\\

\subsection{Manipulation offline eval prompt}

\textbf{Direct}

\vspace{0.3em}
\noindent\fbox{\parbox{0.97\linewidth}{\footnotesize{\texttt{
Summary: The arrows are actions the robot can take.
Red means move the arm forward (away from the camera), blue means move the arm backwards (towards the camera).
Smaller circles are further from the camera and thus move the arm forward, larger circles are closer and thus move the arm backwards.
Do not output anything else, direct answer ith the format, Arrow: [<number>, <number>, etc.].
IMG,
Task: What are the best arrows for the robot follow to pick white coat hanger?
}}}}\\

\textbf{CoT}

\vspace{0.3em}
\noindent\fbox{\parbox{0.97\linewidth}{\footnotesize{\texttt{{
Summary: The arrows are actions the robot can take.
Reason through the task first and at the end summarize the correct action choice(s) with the format, Arrow: [<number>, <number>, etc.].
Description: The robot can only grasp or move objects if the gripper is around the object and closed on the object.
Red means move the arm forward (away from the camera), blue means move the arm backwards (towards the camera).
Smaller circles are further from the camera and thus move the arm forward, larger circles are closer and thus move the arm backwards.
You must include this summarization.
IMG,
Task: What are the best arrows for the robot follow to pick catnip toy?
}}}}}\\

\textbf{Few-shot Direct}

\vspace{0.3em}
\noindent\fbox{\parbox{0.97\linewidth}{\footnotesize{\texttt{{
Summary: (same as above)
IMG,
Task: Erase the writing on the whiteboard.
Arrow: [5, 10],
IMG,
Task: Pick up the iced coffee can.
Arrow: [1],
IMG,
Task: Pick up the string cheese.
Arrow: [8, 15, 3, 13],
IMG,
Task: pick white coat hanger.
}}}}}\\

\textbf{Few-shot CoT}

\vspace{0.3em}
\noindent\fbox{\parbox{0.97\linewidth}{\footnotesize{\texttt{{
Summary: (same as above)
IMG,
Task: Erase the writing on the whiteboard.
The robot is holding an eraser, so it should move it over the marker on the whiteboard. The following arrows look promising:
5. This arrow moves the eraser over the writing and away from the camera and thus towards the whiteboard.
10. This arrow too moves the eraser over the writing and has an even smaller circle (and more red) and thus more towards the whiteboard.
Arrow: [5, 10],
IMG,
Task: ...
Arrow: [5, 10],
IMG,
Task: ...
Arrow: [8, 15, 3, 13],
IMG,
Task: pick oreo.
}}}}}\\

%% file: main.bbl
\begin{thebibliography}{66}
\providecommand{\natexlab}[1]{#1}
\providecommand{\url}[1]{\texttt{#1}}
\expandafter\ifx\csname urlstyle\endcsname\relax
  \providecommand{\doi}[1]{doi: #1}\else
  \providecommand{\doi}{doi: \begingroup \urlstyle{rm}\Url}\fi

\bibitem[Ahn et~al.(2022)Ahn, Brohan, Brown, Chebotar, Cortes, David, Finn, Fu,
  Gopalakrishnan, Hausman, et~al.]{ahn2022can}
Ahn, M., Brohan, A., Brown, N., Chebotar, Y., Cortes, O., David, B., Finn, C.,
  Fu, C., Gopalakrishnan, K., Hausman, K., et~al.
\newblock Do as i can, not as i say: Grounding language in robotic affordances.
\newblock \emph{arXiv preprint arXiv:2204.01691}, 2022.

\bibitem[Alayrac et~al.(2022)Alayrac, Donahue, Luc, Miech, Barr, Hasson, Lenc,
  Mensch, Millican, Reynolds, et~al.]{alayrac2022flamingo}
Alayrac, J.-B., Donahue, J., Luc, P., Miech, A., Barr, I., Hasson, Y., Lenc,
  K., Mensch, A., Millican, K., Reynolds, M., et~al.
\newblock Flamingo: a visual language model for few-shot learning.
\newblock \emph{Advances in Neural Information Processing Systems},
  35:\penalty0 23716--23736, 2022.

\bibitem[Austin et~al.(2021)Austin, Odena, Nye, Bosma, Michalewski, Dohan,
  Jiang, Cai, Terry, Le, et~al.]{austin2021program}
Austin, J., Odena, A., Nye, M., Bosma, M., Michalewski, H., Dohan, D., Jiang,
  E., Cai, C., Terry, M., Le, Q., et~al.
\newblock Program synthesis with large language models.
\newblock \emph{arXiv preprint arXiv:2108.07732}, 2021.

\bibitem[Brohan et~al.(2023)Brohan, Brown, Carbajal, Chebotar, Chen,
  Choromanski, Ding, Driess, Dubey, Finn, et~al.]{brohan2023rt}
Brohan, A., Brown, N., Carbajal, J., Chebotar, Y., Chen, X., Choromanski, K.,
  Ding, T., Driess, D., Dubey, A., Finn, C., et~al.
\newblock Rt-2: Vision-language-action models transfer web knowledge to robotic
  control.
\newblock \emph{arXiv preprint arXiv:2307.15818}, 2023.

\bibitem[Brown et~al.(2020)Brown, Mann, Ryder, Subbiah, Kaplan, Dhariwal,
  Neelakantan, Shyam, Sastry, Askell, et~al.]{brown2020language}
Brown, T., Mann, B., Ryder, N., Subbiah, M., Kaplan, J.~D., Dhariwal, P.,
  Neelakantan, A., Shyam, P., Sastry, G., Askell, A., et~al.
\newblock Language models are few-shot learners.
\newblock \emph{Advances in neural information processing systems},
  33:\penalty0 1877--1901, 2020.

\bibitem[Cai et~al.(2023)Cai, Liu, Mustikovela, Meyer, Chai, Park, and
  Lee]{cai2023making}
Cai, M., Liu, H., Mustikovela, S.~K., Meyer, G.~P., Chai, Y., Park, D., and
  Lee, Y.~J.
\newblock Making large multimodal models understand arbitrary visual prompts.
\newblock \emph{arXiv preprint arXiv:2312.00784}, 2023.

\bibitem[Chen et~al.(2023{\natexlab{a}})Chen, Xia, Ichter, Rao, Gopalakrishnan,
  Ryoo, Stone, and Kappler]{chen2023open}
Chen, B., Xia, F., Ichter, B., Rao, K., Gopalakrishnan, K., Ryoo, M.~S., Stone,
  A., and Kappler, D.
\newblock Open-vocabulary queryable scene representations for real world
  planning.
\newblock In \emph{2023 IEEE International Conference on Robotics and
  Automation (ICRA)}, pp.\  11509--11522. IEEE, 2023{\natexlab{a}}.

\bibitem[Chen et~al.(2024)Chen, Xu, Kirmani, Ichter, Driess, Florence, Sadigh,
  Guibas, and Xia]{chen2024spatialvlm}
Chen, B., Xu, Z., Kirmani, S., Ichter, B., Driess, D., Florence, P., Sadigh,
  D., Guibas, L., and Xia, F.
\newblock Spatialvlm: Endowing vision-language models with spatial reasoning
  capabilities.
\newblock \emph{arXiv preprint arXiv:2401.12168}, 2024.

\bibitem[Chen et~al.(2023{\natexlab{b}})Chen, Djolonga, Padlewski, Mustafa,
  Changpinyo, Wu, Ruiz, Goodman, Wang, Tay, et~al.]{chen2023pali}
Chen, X., Djolonga, J., Padlewski, P., Mustafa, B., Changpinyo, S., Wu, J.,
  Ruiz, C.~R., Goodman, S., Wang, X., Tay, Y., et~al.
\newblock Pali-x: On scaling up a multilingual vision and language model.
\newblock \emph{arXiv preprint arXiv:2305.18565}, 2023{\natexlab{b}}.

\bibitem[Cui et~al.(2022)Cui, Niekum, Gupta, Kumar, and Rajeswaran]{cui2022can}
Cui, Y., Niekum, S., Gupta, A., Kumar, V., and Rajeswaran, A.
\newblock Can foundation models perform zero-shot task specification for robot
  manipulation?
\newblock In \emph{Learning for Dynamics and Control Conference}, pp.\
  893--905. PMLR, 2022.

\bibitem[De~Boer et~al.(2005)De~Boer, Kroese, Mannor, and
  Rubinstein]{de2005tutorial}
De~Boer, P.-T., Kroese, D.~P., Mannor, S., and Rubinstein, R.~Y.
\newblock A tutorial on the cross-entropy method.
\newblock \emph{Annals of operations research}, 134:\penalty0 19--67, 2005.

\bibitem[Dorbala et~al.(2022)Dorbala, Sigurdsson, Piramuthu, Thomason, and
  Sukhatme]{dorbala2022clip}
Dorbala, V.~S., Sigurdsson, G., Piramuthu, R., Thomason, J., and Sukhatme,
  G.~S.
\newblock Clip-nav: Using clip for zero-shot vision-and-language navigation.
\newblock \emph{arXiv preprint arXiv:2211.16649}, 2022.

\bibitem[Firoozi et~al.(2023)Firoozi, Tucker, Tian, Majumdar, Sun, Liu, Zhu,
  Song, Kapoor, Hausman, et~al.]{firoozi2023foundation}
Firoozi, R., Tucker, J., Tian, S., Majumdar, A., Sun, J., Liu, W., Zhu, Y.,
  Song, S., Kapoor, A., Hausman, K., et~al.
\newblock Foundation models in robotics: Applications, challenges, and the
  future.
\newblock \emph{arXiv preprint arXiv:2312.07843}, 2023.

\bibitem[Gadre et~al.(2023)Gadre, Wortsman, Ilharco, Schmidt, and
  Song]{gadre2023cows}
Gadre, S.~Y., Wortsman, M., Ilharco, G., Schmidt, L., and Song, S.
\newblock Cows on pasture: Baselines and benchmarks for language-driven
  zero-shot object navigation.
\newblock In \emph{Proceedings of the IEEE/CVF Conference on Computer Vision
  and Pattern Recognition}, pp.\  23171--23181, 2023.

\bibitem[Gao et~al.(2023)Gao, Sarkar, Xia, Xiao, Wu, Ichter, Majumdar, and
  Sadigh]{gao2023physically}
Gao, J., Sarkar, B., Xia, F., Xiao, T., Wu, J., Ichter, B., Majumdar, A., and
  Sadigh, D.
\newblock Physically grounded vision-language models for robotic manipulation.
\newblock \emph{arXiv preprint arXiv:2309.02561}, 2023.

\bibitem[Gemini et~al.(2023)Gemini, Anil, Borgeaud, Wu, Alayrac, Yu, Soricut,
  Schalkwyk, Dai, Hauth, et~al.]{team2023gemini}
Gemini, T., Anil, R., Borgeaud, S., Wu, Y., Alayrac, J.-B., Yu, J., Soricut,
  R., Schalkwyk, J., Dai, A.~M., Hauth, A., et~al.
\newblock Gemini: a family of highly capable multimodal models.
\newblock \emph{arXiv preprint arXiv:2312.11805}, 2023.

\bibitem[Gemini~Team(2023)]{gemini2023gemini}
Gemini~Team, G.
\newblock Gemini: A family of highly capable multimodal models.
\newblock Technical report, Google, 2023.
\newblock URL
  \url{https://storage.googleapis.com/deepmind-media/gemini/gemini_1_report.pdf}.

\bibitem[Gu et~al.(2023)Gu, Kirmani, Wohlhart, Lu, Arenas, Rao, Yu, Fu,
  Gopalakrishnan, Xu, et~al.]{gu2023rt}
Gu, J., Kirmani, S., Wohlhart, P., Lu, Y., Arenas, M.~G., Rao, K., Yu, W., Fu,
  C., Gopalakrishnan, K., Xu, Z., et~al.
\newblock Rt-trajectory: Robotic task generalization via hindsight trajectory
  sketches.
\newblock \emph{arXiv preprint arXiv:2311.01977}, 2023.

\bibitem[Hu et~al.(2023)Hu, Xie, Jain, Francis, Patrikar, Keetha, Kim, Xie,
  Zhang, Zhao, et~al.]{hu2023toward}
Hu, Y., Xie, Q., Jain, V., Francis, J., Patrikar, J., Keetha, N., Kim, S., Xie,
  Y., Zhang, T., Zhao, Z., et~al.
\newblock Toward general-purpose robots via foundation models: A survey and
  meta-analysis.
\newblock \emph{arXiv preprint arXiv:2312.08782}, 2023.

\bibitem[Huang et~al.(2023{\natexlab{a}})Huang, Mees, Zeng, and
  Burgard]{huang2023visual}
Huang, C., Mees, O., Zeng, A., and Burgard, W.
\newblock Visual language maps for robot navigation.
\newblock In \emph{2023 IEEE International Conference on Robotics and
  Automation (ICRA)}, pp.\  10608--10615. IEEE, 2023{\natexlab{a}}.

\bibitem[Huang et~al.(2022{\natexlab{a}})Huang, Abbeel, Pathak, and
  Mordatch]{huang2022language}
Huang, W., Abbeel, P., Pathak, D., and Mordatch, I.
\newblock Language models as zero-shot planners: Extracting actionable
  knowledge for embodied agents.
\newblock In \emph{International Conference on Machine Learning}, pp.\
  9118--9147. PMLR, 2022{\natexlab{a}}.

\bibitem[Huang et~al.(2022{\natexlab{b}})Huang, Xia, Xiao, Chan, Liang,
  Florence, Zeng, Tompson, Mordatch, Chebotar, et~al.]{huang2022inner}
Huang, W., Xia, F., Xiao, T., Chan, H., Liang, J., Florence, P., Zeng, A.,
  Tompson, J., Mordatch, I., Chebotar, Y., et~al.
\newblock Inner monologue: Embodied reasoning through planning with language
  models.
\newblock \emph{arXiv preprint arXiv:2207.05608}, 2022{\natexlab{b}}.

\bibitem[Huang et~al.(2023{\natexlab{b}})Huang, Wang, Zhang, Li, Wu, and
  Fei-Fei]{huang2023voxposer}
Huang, W., Wang, C., Zhang, R., Li, Y., Wu, J., and Fei-Fei, L.
\newblock Voxposer: Composable 3d value maps for robotic manipulation with
  language models.
\newblock \emph{arXiv preprint arXiv:2307.05973}, 2023{\natexlab{b}}.

\bibitem[Itseez(2015)]{itseez2015opencv}
Itseez.
\newblock Open source computer vision library.
\newblock \url{https://github.com/itseez/opencv}, 2015.

\bibitem[Jiang et~al.(2022)Jiang, Gupta, Zhang, Wang, Dou, Chen, Fei-Fei,
  Anandkumar, Zhu, and Fan]{jiang2022vima}
Jiang, Y., Gupta, A., Zhang, Z., Wang, G., Dou, Y., Chen, Y., Fei-Fei, L.,
  Anandkumar, A., Zhu, Y., and Fan, L.
\newblock Vima: General robot manipulation with multimodal prompts.
\newblock \emph{arXiv}, 2022.

\bibitem[Koh et~al.(2024)Koh, Lo, Jang, Duvvur, Lim, Huang, Neubig, Zhou,
  Salakhutdinov, and Fried]{koh2024visualwebarena}
Koh, J.~Y., Lo, R., Jang, L., Duvvur, V., Lim, M.~C., Huang, P.-Y., Neubig, G.,
  Zhou, S., Salakhutdinov, R., and Fried, D.
\newblock Visualwebarena: Evaluating multimodal agents on realistic visual web
  tasks.
\newblock \emph{arXiv preprint arXiv:2401.13649}, 2024.

\bibitem[Kojima et~al.(2022)Kojima, Gu, Reid, Matsuo, and
  Iwasawa]{kojima2022large}
Kojima, T., Gu, S.~S., Reid, M., Matsuo, Y., and Iwasawa, Y.
\newblock Large language models are zero-shot reasoners.
\newblock \emph{Advances in neural information processing systems},
  35:\penalty0 22199--22213, 2022.

\bibitem[Lester et~al.(2021)Lester, Al-Rfou, and Constant]{lester2021power}
Lester, B., Al-Rfou, R., and Constant, N.
\newblock The power of scale for parameter-efficient prompt tuning.
\newblock \emph{arXiv preprint arXiv:2104.08691}, 2021.

\bibitem[Li \& Liang(2021)Li and Liang]{li2021prefix}
Li, X.~L. and Liang, P.
\newblock Prefix-tuning: Optimizing continuous prompts for generation.
\newblock \emph{arXiv preprint arXiv:2101.00190}, 2021.

\bibitem[Liang et~al.(2023)Liang, Huang, Xia, Xu, Hausman, Ichter, Florence,
  and Zeng]{liang2023code}
Liang, J., Huang, W., Xia, F., Xu, P., Hausman, K., Ichter, B., Florence, P.,
  and Zeng, A.
\newblock Code as policies: Language model programs for embodied control.
\newblock In \emph{2023 IEEE International Conference on Robotics and
  Automation (ICRA)}, pp.\  9493--9500. IEEE, 2023.

\bibitem[Lin et~al.(2023)Lin, Agia, Migimatsu, Pavone, and
  Bohg]{lin2023text2motion}
Lin, K., Agia, C., Migimatsu, T., Pavone, M., and Bohg, J.
\newblock Text2motion: From natural language instructions to feasible plans.
\newblock \emph{arXiv preprint arXiv:2303.12153}, 2023.

\bibitem[Liu et~al.(2023{\natexlab{a}})Liu, Jiang, Zhang, Liu, Zhang, Biswas,
  and Stone]{liu2023llm+}
Liu, B., Jiang, Y., Zhang, X., Liu, Q., Zhang, S., Biswas, J., and Stone, P.
\newblock Llm+ p: Empowering large language models with optimal planning
  proficiency.
\newblock \emph{arXiv preprint arXiv:2304.11477}, 2023{\natexlab{a}}.

\bibitem[Liu et~al.(2023{\natexlab{b}})Liu, Dong, Zhang, Luo, Gao, Huang, Gong,
  and Wang]{liu20233daxiesprompts}
Liu, D., Dong, X., Zhang, R., Luo, X., Gao, P., Huang, X., Gong, Y., and Wang,
  Z.
\newblock 3daxiesprompts: Unleashing the 3d spatial task capabilities of
  gpt-4v.
\newblock \emph{arXiv preprint arXiv:2312.09738}, 2023{\natexlab{b}}.

\bibitem[Liu et~al.(2023{\natexlab{c}})Liu, Bahety, and Song]{liu2023reflect}
Liu, Z., Bahety, A., and Song, S.
\newblock Reflect: Summarizing robot experiences for failure explanation and
  correction.
\newblock \emph{arXiv preprint arXiv:2306.15724}, 2023{\natexlab{c}}.

\bibitem[Ma et~al.(2023)Ma, Liang, Wang, Huang, Bastani, Jayaraman, Zhu, Fan,
  and Anandkumar]{ma2023eureka}
Ma, Y.~J., Liang, W., Wang, G., Huang, D.-A., Bastani, O., Jayaraman, D., Zhu,
  Y., Fan, L., and Anandkumar, A.
\newblock Eureka: Human-level reward design via coding large language models.
\newblock \emph{arXiv preprint arXiv:2310.12931}, 2023.

\bibitem[Mirchandani et~al.(2023)Mirchandani, Xia, Florence, Ichter, Driess,
  Arenas, Rao, Sadigh, and Zeng]{mirchandani2023large}
Mirchandani, S., Xia, F., Florence, P., Ichter, B., Driess, D., Arenas, M.~G.,
  Rao, K., Sadigh, D., and Zeng, A.
\newblock Large language models as general pattern machines.
\newblock \emph{arXiv preprint arXiv:2307.04721}, 2023.

\bibitem[OpenAI(2023)]{openai2023gpt4v}
OpenAI.
\newblock Gpt-4v(ision) system card.
\newblock Technical report, OpenAI, 2023.
\newblock URL \url{https://openai.com/research/gpt-4v-system-card}.

\bibitem[Padalkar et~al.(2023)Padalkar, Pooley, Jain, Bewley, Herzog, Irpan,
  Khazatsky, Rai, Singh, Brohan, et~al.]{padalkar2023open}
Padalkar, A., Pooley, A., Jain, A., Bewley, A., Herzog, A., Irpan, A.,
  Khazatsky, A., Rai, A., Singh, A., Brohan, A., et~al.
\newblock Open x-embodiment: Robotic learning datasets and rt-x models.
\newblock \emph{arXiv preprint arXiv:2310.08864}, 2023.

\bibitem[Pryzant et~al.(2023)Pryzant, Iter, Li, Lee, Zhu, and
  Zeng]{pryzant2023automatic}
Pryzant, R., Iter, D., Li, J., Lee, Y.~T., Zhu, C., and Zeng, M.
\newblock Automatic prompt optimization with" gradient descent" and beam
  search.
\newblock \emph{arXiv preprint arXiv:2305.03495}, 2023.

\bibitem[Radford et~al.(2021)Radford, Kim, Hallacy, Ramesh, Goh, Agarwal,
  Sastry, Askell, Mishkin, Clark, et~al.]{radford2021learning}
Radford, A., Kim, J.~W., Hallacy, C., Ramesh, A., Goh, G., Agarwal, S., Sastry,
  G., Askell, A., Mishkin, P., Clark, J., et~al.
\newblock Learning transferable visual models from natural language
  supervision.
\newblock In \emph{International conference on machine learning}, pp.\
  8748--8763. PMLR, 2021.

\bibitem[Raman et~al.(2022)Raman, Cohen, Rosen, Idrees, Paulius, and
  Tellex]{raman2022planning}
Raman, S.~S., Cohen, V., Rosen, E., Idrees, I., Paulius, D., and Tellex, S.
\newblock Planning with large language models via corrective re-prompting.
\newblock In \emph{NeurIPS 2022 Foundation Models for Decision Making
  Workshop}, 2022.

\bibitem[Reed et~al.(2022)Reed, Zolna, Parisotto, Colmenarejo, Novikov,
  Barth-Maron, Gimenez, Sulsky, Kay, Springenberg, et~al.]{reed2022generalist}
Reed, S., Zolna, K., Parisotto, E., Colmenarejo, S.~G., Novikov, A.,
  Barth-Maron, G., Gimenez, M., Sulsky, Y., Kay, J., Springenberg, J.~T.,
  et~al.
\newblock A generalist agent.
\newblock \emph{arXiv preprint arXiv:2205.06175}, 2022.

\bibitem[Shah et~al.(2023{\natexlab{a}})Shah, Equi, Osi{\'n}ski, Xia, Ichter,
  and Levine]{shah2023navigation}
Shah, D., Equi, M.~R., Osi{\'n}ski, B., Xia, F., Ichter, B., and Levine, S.
\newblock Navigation with large language models: Semantic guesswork as a
  heuristic for planning.
\newblock In \emph{Conference on Robot Learning}, pp.\  2683--2699. PMLR,
  2023{\natexlab{a}}.

\bibitem[Shah et~al.(2023{\natexlab{b}})Shah, Osi{\'n}ski, Levine,
  et~al.]{shah2023lm}
Shah, D., Osi{\'n}ski, B., Levine, S., et~al.
\newblock Lm-nav: Robotic navigation with large pre-trained models of language,
  vision, and action.
\newblock In \emph{Conference on Robot Learning}, pp.\  492--504. PMLR,
  2023{\natexlab{b}}.

\bibitem[Shridhar et~al.(2022)Shridhar, Manuelli, and Fox]{shridhar2022cliport}
Shridhar, M., Manuelli, L., and Fox, D.
\newblock Cliport: What and where pathways for robotic manipulation.
\newblock In \emph{Conference on Robot Learning}, pp.\  894--906. PMLR, 2022.

\bibitem[Shtedritski et~al.(2023)Shtedritski, Rupprecht, and
  Vedaldi]{shtedritski2023does}
Shtedritski, A., Rupprecht, C., and Vedaldi, A.
\newblock What does clip know about a red circle? visual prompt engineering for
  vlms.
\newblock \emph{arXiv preprint arXiv:2304.06712}, 2023.

\bibitem[Silver et~al.(2023)Silver, Dan, Srinivas, Tenenbaum, Kaelbling, and
  Katz]{silver2023generalized}
Silver, T., Dan, S., Srinivas, K., Tenenbaum, J.~B., Kaelbling, L.~P., and
  Katz, M.
\newblock Generalized planning in pddl domains with pretrained large language
  models.
\newblock \emph{arXiv preprint arXiv:2305.11014}, 2023.

\bibitem[Singh et~al.(2023)Singh, Blukis, Mousavian, Goyal, Xu, Tremblay, Fox,
  Thomason, and Garg]{singh2023progprompt}
Singh, I., Blukis, V., Mousavian, A., Goyal, A., Xu, D., Tremblay, J., Fox, D.,
  Thomason, J., and Garg, A.
\newblock Progprompt: Generating situated robot task plans using large language
  models.
\newblock In \emph{2023 IEEE International Conference on Robotics and
  Automation (ICRA)}, pp.\  11523--11530. IEEE, 2023.

\bibitem[Wang et~al.(2023{\natexlab{a}})Wang, Xie, Jiang, Mandlekar, Xiao, Zhu,
  Fan, and Anandkumar]{wang2023voyager}
Wang, G., Xie, Y., Jiang, Y., Mandlekar, A., Xiao, C., Zhu, Y., Fan, L., and
  Anandkumar, A.
\newblock Voyager: An open-ended embodied agent with large language models.
\newblock \emph{arXiv preprint arXiv:2305.16291}, 2023{\natexlab{a}}.

\bibitem[Wang et~al.(2023{\natexlab{b}})Wang, Zhang, Chen, and
  Sreenath]{wang2023prompt}
Wang, Y.-J., Zhang, B., Chen, J., and Sreenath, K.
\newblock Prompt a robot to walk with large language models.
\newblock \emph{arXiv preprint arXiv:2309.09969}, 2023{\natexlab{b}}.

\bibitem[Wang et~al.(2023{\natexlab{c}})Wang, Cai, Liu, Ma, and
  Liang]{wang2023describe}
Wang, Z., Cai, S., Liu, A., Ma, X., and Liang, Y.
\newblock Describe, explain, plan and select: Interactive planning with large
  language models enables open-world multi-task agents.
\newblock \emph{arXiv preprint arXiv:2302.01560}, 2023{\natexlab{c}}.

\bibitem[Wei et~al.(2022)Wei, Wang, Schuurmans, Bosma, Xia, Chi, Le, Zhou,
  et~al.]{wei2022chain}
Wei, J., Wang, X., Schuurmans, D., Bosma, M., Xia, F., Chi, E., Le, Q.~V.,
  Zhou, D., et~al.
\newblock Chain-of-thought prompting elicits reasoning in large language
  models.
\newblock \emph{Advances in Neural Information Processing Systems},
  35:\penalty0 24824--24837, 2022.

\bibitem[Wen et~al.(2023)Wen, Yang, Fu, Wang, Cai, Li, Ma, Li, Xu, Shang,
  et~al.]{wen2023road}
Wen, L., Yang, X., Fu, D., Wang, X., Cai, P., Li, X., Ma, T., Li, Y., Xu, L.,
  Shang, D., et~al.
\newblock On the road with gpt-4v (ision): Early explorations of
  visual-language model on autonomous driving.
\newblock \emph{arXiv preprint arXiv:2311.05332}, 2023.

\bibitem[Wu et~al.(2023)Wu, Antonova, Kan, Lepert, Zeng, Song, Bohg,
  Rusinkiewicz, and Funkhouser]{wu2023tidybot}
Wu, J., Antonova, R., Kan, A., Lepert, M., Zeng, A., Song, S., Bohg, J.,
  Rusinkiewicz, S., and Funkhouser, T.
\newblock Tidybot: Personalized robot assistance with large language models.
\newblock \emph{arXiv preprint arXiv:2305.05658}, 2023.

\bibitem[Xu et~al.(2022)Xu, Chen, Du, Shao, Wang, Li, and Yang]{xu2022gps}
Xu, H., Chen, Y., Du, Y., Shao, N., Wang, Y., Li, H., and Yang, Z.
\newblock Gps: Genetic prompt search for efficient few-shot learning.
\newblock \emph{arXiv preprint arXiv:2210.17041}, 2022.

\bibitem[Xu et~al.(2023)Xu, Zhou, Yan, Gu, Arnab, Sun, Wang, and
  Schmid]{xu2023pixel}
Xu, J., Zhou, X., Yan, S., Gu, X., Arnab, A., Sun, C., Wang, X., and Schmid, C.
\newblock Pixel aligned language models.
\newblock \emph{arXiv preprint arXiv:2312.09237}, 2023.

\bibitem[Yan et~al.(2023)Yan, Yang, Zhu, Lin, Li, Wang, Yang, Zhong, McAuley,
  Gao, et~al.]{yan2023gpt}
Yan, A., Yang, Z., Zhu, W., Lin, K., Li, L., Wang, J., Yang, J., Zhong, Y.,
  McAuley, J., Gao, J., et~al.
\newblock Gpt-4v in wonderland: Large multimodal models for zero-shot
  smartphone gui navigation.
\newblock \emph{arXiv preprint arXiv:2311.07562}, 2023.

\bibitem[Yang et~al.(2023{\natexlab{a}})Yang, Wang, Lu, Liu, Le, Zhou, and
  Chen]{yang2023large}
Yang, C., Wang, X., Lu, Y., Liu, H., Le, Q.~V., Zhou, D., and Chen, X.
\newblock Large language models as optimizers.
\newblock \emph{arXiv preprint arXiv:2309.03409}, 2023{\natexlab{a}}.

\bibitem[Yang et~al.(2023{\natexlab{b}})Yang, Zhang, Li, Zou, Li, and
  Gao]{yang2023set}
Yang, J., Zhang, H., Li, F., Zou, X., Li, C., and Gao, J.
\newblock Set-of-mark prompting unleashes extraordinary visual grounding in
  gpt-4v.
\newblock \emph{arXiv preprint arXiv:2310.11441}, 2023{\natexlab{b}}.

\bibitem[Yang et~al.(2023{\natexlab{c}})Yang, Li, Lin, Wang, Lin, Liu, and
  Wang]{yang2023dawn}
Yang, Z., Li, L., Lin, K., Wang, J., Lin, C.-C., Liu, Z., and Wang, L.
\newblock The dawn of lmms: Preliminary explorations with gpt-4v (ision).
\newblock \emph{arXiv preprint arXiv:2309.17421}, 9\penalty0 (1):\penalty0 1,
  2023{\natexlab{c}}.

\bibitem[Yu et~al.(2016)Yu, Poirson, Yang, Berg, and Berg]{yu2016modeling}
Yu, L., Poirson, P., Yang, S., Berg, A.~C., and Berg, T.~L.
\newblock Modeling context in referring expressions.
\newblock In \emph{Computer Vision--ECCV 2016: 14th European Conference,
  Amsterdam, The Netherlands, October 11-14, 2016, Proceedings, Part II 14},
  pp.\  69--85. Springer, 2016.

\bibitem[Yu et~al.(2023)Yu, Gileadi, Fu, Kirmani, Lee, Arenas, Chiang, Erez,
  Hasenclever, Humplik, et~al.]{yu2023language}
Yu, W., Gileadi, N., Fu, C., Kirmani, S., Lee, K.-H., Arenas, M.~G., Chiang,
  H.-T.~L., Erez, T., Hasenclever, L., Humplik, J., et~al.
\newblock Language to rewards for robotic skill synthesis.
\newblock \emph{arXiv preprint arXiv:2306.08647}, 2023.

\bibitem[Zeng et~al.(2021)Zeng, Florence, Tompson, Welker, Chien, Attarian,
  Armstrong, Krasin, Duong, Sindhwani, et~al.]{zeng2021transporter}
Zeng, A., Florence, P., Tompson, J., Welker, S., Chien, J., Attarian, M.,
  Armstrong, T., Krasin, I., Duong, D., Sindhwani, V., et~al.
\newblock Transporter networks: Rearranging the visual world for robotic
  manipulation.
\newblock In \emph{Conference on Robot Learning}, pp.\  726--747. PMLR, 2021.

\bibitem[Zeng et~al.(2022)Zeng, Attarian, Ichter, Choromanski, Wong, Welker,
  Tombari, Purohit, Ryoo, Sindhwani, et~al.]{zeng2022socratic}
Zeng, A., Attarian, M., Ichter, B., Choromanski, K., Wong, A., Welker, S.,
  Tombari, F., Purohit, A., Ryoo, M., Sindhwani, V., et~al.
\newblock Socratic models: Composing zero-shot multimodal reasoning with
  language.
\newblock \emph{arXiv preprint arXiv:2204.00598}, 2022.

\bibitem[Zheng et~al.(2024)Zheng, Gou, Kil, Sun, and Su]{zheng2024gpt}
Zheng, B., Gou, B., Kil, J., Sun, H., and Su, Y.
\newblock Gpt-4v (ision) is a generalist web agent, if grounded.
\newblock \emph{arXiv preprint arXiv:2401.01614}, 2024.

\bibitem[Zhou et~al.(2022)Zhou, Muresanu, Han, Paster, Pitis, Chan, and
  Ba]{zhou2022large}
Zhou, Y., Muresanu, A.~I., Han, Z., Paster, K., Pitis, S., Chan, H., and Ba, J.
\newblock Large language models are human-level prompt engineers.
\newblock \emph{arXiv preprint arXiv:2211.01910}, 2022.

\end{thebibliography}
